\pdfoutput=1
\documentclass[11pt]{article}

\usepackage{acl}

\usepackage{times}
\usepackage{latexsym}
\usepackage{hyperref}

\usepackage[T1]{fontenc}
\usepackage[utf8]{inputenc}

\usepackage{microtype}

\usepackage{booktabs}

\usepackage{graphics}

\usepackage{amsmath}
\usepackage{multirow, hhline, graphicx, subcaption, helvet}
\mathchardef\mhyphen="2D
\newenvironment{itemizesquish}{\begin{list}{\labelitemi}{\setlength{\itemsep}{-0.2em}\setlength{\labelwidth}{0.5em}\setlength{\leftmargin}{\labelwidth}\addtolength{\leftmargin}{\labelsep}}}{\end{list}}
\newenvironment{enumeratesquish}{\begin{list}{\addtocounter{enumi}{1}\labelenumi}{\setlength{\itemsep}{0em}\setlength{\labelwidth}{0.5em}\setlength{\leftmargin}{\labelwidth}\addtolength{\leftmargin}{\labelsep}}}{\end{list}\setcounter{enumi}{0}}

\usepackage{microtype}

\usepackage{todonotes}
\usepackage{algorithm}
\usepackage{algpseudocode}
\algnewcommand\algorithmicyield{\textbf{yield}}
\algnewcommand\Yield{\algorithmicyield{} }

\algnewcommand{\LeftComment}[1]{\Statex \(\triangleright\) #1}

\usepackage{tabularx}
\usepackage{makecell}
\usepackage{array}
\newcolumntype{L}{>{\RaggedRight\arraybackslash}X}
\newcolumntype{C}[1]{>{\centering\arraybackslash%
\hsize=#1\hsize\linewidth=\hsize}X}

\newcommand{\Note}[2]{} 
\newcommand{\SideNote}[2]{}
\renewcommand{\Note}[2]{\todo[color=#1,size=\small, inline=true]{#2}} 
\renewcommand{\SideNote}[2]{\todo[color=#1,size=\small]{#2}} % 

\newcommand{\Kevinin}[1]{\textcolor{blue}{\bf \small [ #1 --Kevin]}}

\usepackage{xspace}
\newcommand{\cs}{\textsc{base}\xspace}
\newcommand{\css}{\textsc{span}\xspace}
\newcommand{\cssnqt}{\textsc{type}$_s$\xspace}
\newcommand{\cssrqt}{\textsc{type}$_o$\xspace}
\newcommand{\cssrankqt}{\textsc{type}$_r$\xspace}
\newcommand{\qtfamily}{\textsc{type}\xspace}
\newcommand{\human}{\textsc{human}\xspace}
\newcommand{\inquisitive}{\textsc{Inquisitive}\xspace}
\newcommand{\std}[1]{{\tiny $\pm$#1}}

\title{%``How is the climate changing?" 
``What makes a question inquisitive?''\\A Study on Type-Controlled Inquisitive Question Generation
}
 \author{Lingyu Gao\thanks{~~Partially done as part of an internship at Educational Testing Service.}  \textsuperscript{1}, 
  Debanjan Ghosh\textsuperscript{2}, 
  \textbf{and} \textbf{Kevin Gimpel}\textsuperscript{1}\\ 
  \textsuperscript{1}Toyota Technological Institute at Chicago \\
  \textsuperscript{2}Educational Testing Service\\
  {\tt \{lygao, kgimpel\}@ttic.edu}, \\{\tt dghosh@ets.org}
  }

\begin{document}
\maketitle
\begin{abstract}
We propose a type-controlled framework for inquisitive question generation. We annotate an inquisitive question dataset with question types, train question type classifiers, and finetune models for type-controlled question generation. Empirical results demonstrate that we can generate a variety of questions that adhere to specific types while drawing from the source texts. We also investigate strategies for selecting a single question from a generated set, considering both an informative vs.~inquisitive question classifier and a pairwise ranker trained from a small set of expert annotations. Question selection using the pairwise ranker yields strong results in automatic and manual evaluation. Our human evaluation assesses multiple aspects of the generated questions, finding that the ranker chooses questions with the best syntax (4.59), semantics (4.37), and inquisitiveness (3.92) on a scale of 1-5, even rivaling the performance of human-written questions. 

%\Kevinin{I added semantics here as well, since \cssrankqt is also best in semantics (among non-\human questions.. but in terms of inquisitiveness \cssrankqt is also not as good as \human, so I assumed we were talking about non-human questions in this context anyway.}

% that we collected. 
%Empirical results demonstrate: (1) we can generate a variety of questions that adhere to specific types while drawing from 
%following the types conditioned on the question types based on 
%the source text; (2) question selection using the pairwise ranker 
%the model where question types are selected via the pairwise ranking classifier

%Question generation is traditionally to generate questions that could be answered by given context. However, human tend to ask curiosity-driven questions that seeking deeper information when reading documents, such as background information, explanations for certain concepts, etc. 
%While the inquisitive questions could be more novel and diverse without limitations by answers, this new task would add to the difficulty for auto-generation and evaluation. 
%In this research, we classify the inquisitive questions to different question types, and apply them as control codes for question generation with transformer-based architectures.
%We demonstrate that our generations are novel as well as controllable, with amplified paraphrase forms, also are of good quality that is close or even better than human behaviours.
\end{abstract}

\section{Introduction} \label{section:intro}

%\Kevin{I created latex commands for model names and the inquisitive dataset so that we can get uniformity in the text and more easily make changes in the future. I think the current naming scheme can be improved, but it's a starting point (and it will be much easier to experiment with changing it once we are using the appropriate latex commands whenever we want to mention them)}

Recently, interest has grown in the task of automatic question generation (AQG) from text \cite{sun-etal-2018-answer, kumar2020clarq}.  AQG is useful in building conversational AI systems \cite{bordes2016learning, gao-etal-2019-interconnected}, generating synthetic examples for QA \cite{alberti-etal-2019-synthetic, 10.5555/3454287.3455457, sultan-etal-2020-importance}, and 
%most importantly - 
educational applications, such as intelligent tutoring and instructional games \cite{ICWSM18LearningQ, flor-riordan-2018-semantic}. In the majority of such studies, AQG focuses on generating factual questions that tend to ask about specific information %appear 
in the text (i.e., ``who did what to whom'') \cite{du-etal-2017-learning}.

\begin{table}[t]
\centering
\small
\setlength{\tabcolsep}{6pt}
\begin{tabular}{@{} p{1.4cm}p{5.8cm} @{}}
\toprule
Context &  \dots The plan places an indicated value on the real estate operation, Santa Fe Pacific Realty Corp., of \$ 2 billion. \\\midrule
Source \quad sentence & Santa Fe Pacific directors are expected to \textbf{review} the plan at a meeting today, according to people familiar with the transaction. \\\midrule
\cs & What kind of meeting?\\ 
\css & How will the directors review the plan?\\
%cssnqt & what is that? \\
%cssrqt & Why are the directors reviewing the plan at a meeting?\\
%cssrank & Why are they reviewing the plan? \\
\midrule
Explanation & Why are they reviewing the plan? \\ 
Background & Is it expected to review the plan today? \\ 
Elaboration & What will the review entail? \\ 
Instantiation & Which directors are expected to review the plan? \\ 
Definition & what is that? \\ 
Forward & What are the directors expected to review? \\ \midrule
Informative & Who are Santa Fe Pacific directors expected to review? \\
\bottomrule
 % &   {\bf Prior Turn:} Natural selection isn't anything to do with evolution.  \\
 % Disagree &  {\bf Current Turn:} Natural selection is one of the mechanisms of evolution. creationists with limited knowledge can't figure that out. \\
\end{tabular}
\caption{\label{source_exp} Examples of generated questions given the article context and source sentence (with span in \textbf{bold}). %The \cs model is BART finetuned to generate questions from the context and source and the \css model additionally uses the span. The remaining questions are from our controllable generator for the provided question types. 
}
\end{table}
% \Lingyu{cssrqt question are different from Explanation question because of the nucleus sampling method in generation. The reference type is Explanation here.}

Instead of asking factual questions with answers already present in the text, \newcite{ko2020inquisitive} argued that human readers instinctively ask  questions that are curiosity-driven, answer-agnostic, and seek a high-level understanding of the document being read. 
They released a dataset  of such curiosity-driven questions (henceforth \inquisitive; for details, see Section \ref{section:data}). 
The objective of our work is to generate deeper, inquisitive questions based on the \inquisitive dataset. 
%\Kevinin{we can add our motivation here to satisfy R1}

%Our motivation is two-fold. First, 
%Our question generation models (Table~\ref{source_exp}) can generate a variety of questions conditioned on the question types. 
Our motivations for generating inquisitive questions are two-fold. 
%Thus, on one hand, users such as the 
Educators can obtain \emph{diverse} questions for a specific source text when designing quizzes or choosing questions to test students' reading comprehension ability. They can focus into different aspects of the context (e.g., questioning the background information or asking to elaborate a fact) for diverse question generation \cite{cho-etal-2019-mixture,wang-etal-2020-diversify,sultan-etal-2020-importance}. Likewise, students can also be assisted in knowledge acquisition and building reasoning skills by practising over a large number of diverse questions  \cite{cao-wang-2021-controllable}. 
%\Kevinin{moved this para up here per R1's comment; we can add more to this!}

Though our initial efforts are similar to \citet{ko2020inquisitive}, 
%i.e., question generation by concatenating the source sentence with metadata from \inquisitive, 
we found this to be insufficient as it does not leverage the inherent diversity of question types in the dataset. \newcite{ko2020inquisitive} concatenated the context, source sentence, and the question to learn a language model for question generation using GPT-2 \cite{radford2019language}. 
%\Kevinin{to do: add a little more motivation here for why we are doing controlled generation; base/span generate dominant question types.} 
On the contrary, we first annotated 1550 questions from the training partition of the \inquisitive dataset to identify the question types, such as questions requesting background information, asking about the cause of an event, asking for details on underspecified facts, etc. (see Section \ref{subsection:qtypes} for details). We finetune a RoBERTa-large model \cite{liu2019roberta} to predict the question types on the rest of the dataset. We then use the question types in a controlled generation framework based on BART \cite{lewis-etal-2020-bart} to generate type-specific inquisitive questions.  %\cite{keskar2019ctrl,krause-etal-2021-gedi-generative} with BART \cite{lewis-etal-2020-bart},  a large pretrained transformer model to generate the inquisitive questions. 

%On the contrary, we annotated a subset of the \inquisitive questions to a set of question types (see Section \ref{subsection:qannot}) and in turn used the types as control codes \cite{keskar2019ctrl,krause-etal-2021-gedi-generative} to generate inquisitive questions.    

Consider the example in Table~\ref{source_exp}. The \cs model is BART finetuned on \inquisitive to generate questions from the context and source sentence. The \css model additionally uses the span, a part of the source sentence the annotators are curious about. 
%\Kevinin{to do: briefly define span} 
%\Kevinin{For this initial table, we could drop the \cs row, which would streamline the discussion here and save space.}
We then show six questions of specific types (e.g., Explanation, \dots, Forward) generated by our type-controlled finetuned BART model. 
In comparison, the informative question is generated by finetuning on SQuAD \cite{rajpurkar-etal-2016-squad}, a popular dataset for generating factual questions. The informative question asks for surface-level information (``who are Santa Fe Pacific directors expected to review?'') whereas the inquisitive questions ask for deeper information (e.g., ``why are they reviewing the plan?''), such as the reason for the directors' actions.

As mentioned earlier, our motivations for generating diverse inquisitive questions are to provide educational tools and resources. However, there are also cases where an educator or student may prefer only a single high-quality question for a span or a ranked list of questions. We investigate two strategies for automatic question selection/ranking for this latter scenario. 
%Given, the models are trained to generate a variety of questions based on a specific source text (we are using six question types as control codes) we investigate two  scenarios to select the \emph{best} questions. 
The first strategy ranks questions using 
%of generating a question of each type and then a single question is selected automatically using 
an inquisitive vs.~informative question classifier, 
%(denoted as \cssnqt henceforth). 
%as \emph{context-source-span-nqtype} or for brevity, \cssnqt in section \ref{section:method} and section \ref{section:result}, respectively).
where questions from SQuAD are used as  informative questions. 
In the second strategy, we collect expert annotations of partial rankings for a subset of generated questions, and then train a pairwise ranker to select the best question (denoted as \cssrankqt).  %\emph{context-source-span-rank} or, \emph{cssrankqt} in section \ref{section:method} and section \ref{section:result}). 
In automatic evaluation, we find that 
%measures the extent of repeating source text and gold questions into generated questions and observe that questions generated from 
\cssrankqt yields questions that have reasonably strong match to references while also being novel relative to the training set (Section \ref{subsection:autoeval}). 
%\Kevinin{We could comment out the previous sentence to save space as I don't think we want readers to focus on the automatic evaluation}
%Likewise, 
We report a large-scale human evaluation via Mechanical Turk and demonstrate that questions generated from the same \cssrankqt model have the best syntax (4.59), semantics (4.37), and inquisitiveness (3.92) on a scale of 1-5 
%(1 is the lowest and 5 is the highest, 
(Section \ref{subsection:humeval}).
We make the annotations, code, and the MTurk judgements from our research publicly available.\footnote{\scriptsize \url{https://github.com/EducationalTestingService/inquisitive-questions}}

\section{Data} \label{section:data}

We will now describe the annotation of questions with question types, which is one of the main contributions of our work. We describe the annotation process in detail in Section \ref{subsection:qtypes}. But first, we briefly introduce the \inquisitive dataset.%\footnote{Interested readers can consult \citet{ko2020inquisitive} for more information.}

%\Kevinin{Suggestions for structuring this section: it would be best to first say what new contribution we are going to describe in this section, then provide background about \inquisitive (maybe in a separate subsection). }

%Our \emph{training} and \emph{test} data for question generation is based on \inquisitive \cite{ko2020inquisitive}. 
Human annotators created the inquisitive questions while reading the 
initial part (i.e., five sentences) of news articles from the WSJ portion of the Penn Treebank \citep{marcus-etal-1993-building} or Associated Press articles from the TIPSTER corpus \citep{AP_tipster}.\footnote{They also use Newsela \citep{DBLP:journals/tacl/XuCN15} but it was not publicly released.
%\Lingyuin{Maybe we could point it out as they use three datasets in total?}
} 
Annotators first highlighted a span within the sentence that they were curious about and then wrote a maximum of three questions. Next, a separate set of annotators validated the questions and excluded unqualified questions (around 5\%). 

% In order to generate questions that are closer to those proposed by human readers, we adopt INQUISITIVE \cite{ko2020inquisitive} that capture ques-077tions from human annotators during their reading procedures. This dataset is created with two steps. For the first step, the workers are provided with the first 5 sentences of an article sentence by sentence, and they could choose to ask 0 to 3 questions when reading each sentence. If they want to ask a question, they would highlight a span within the sentence they are reading that they are curious about. For each sentence, there are 5 distinct qualified Amazon Mechanical Turk workers working on it. And for the second step, there are 3 distinct workers validating each question proposed in the first step, and excluded the unqualified questions (around 5\%) before the release.

An instance in \inquisitive has the following components: a \textbf{source sentence},  the sentence the annotator read when asking the question, \textbf{context} that includes all the sentences before the source sentence in the same article, a \textbf{span} within the source sentence the annotators were most curious about, and finally, the \textbf{question} the annotator wrote.
%After removal of the erroneous instances \Kevinin{need to say what you mean by ``erroneous instances''} 
\inquisitive is split into \emph{training} (15,897 instances), \emph{test} (1,885 instances), and \emph{dev} (1,984 instances). %\Lingyuin{Do we need to explain that we remove 1 line with empty span in dev set?}     

%After removing the duplicate and empty spans samples, the statistics of datasets are shown in Table~\ref{inq_table}. \Debanjan{is it something you have done or the original paper does too.} \Lingyu{This is something I have done. I don't think the original paper does so.} 

\iffalse 
\begin{table}[t]
\begin{center}
\begin{tabular}{cccc}\hline
 & Train & Dev & Test \\\hline
\#samples & 15896 & 1983 & 1884 \\\hline
\end{tabular}
\end{center}
\caption{\label{inq_table}Statistics of datasets, where `\#' denoting the number of samples.}
\end{table}
\fi 
% We annotate the first 100 samples in the training data with one annotator to see whether we could derive the answer from the source sentence/previous context, and the result is shown in Table~\ref{inq_ans}. Which shows that under most circumstances, the answers are not in the source sentences. \Debanjan{unclear - why we are doing this? how did this fit into this section?} \Lingyu{This part could be removed.}

% \begin{table}[t]
% \begin{center}
% \begin{tabular}{ccccc}\hline
%  & source & context & absent & Total \\\hline
% \#samples & 7 & 16 & 77 & 100 \\\hline
% \end{tabular}
% \end{center}
% \caption{\label{inq_ans} Position where the answer could be derived, where the `source' denotes source sentence, `context' denotes previous context, and `absent' denotes that the answer is not in any of above sentences.}
% \end{table}
\subsection{Question Type Annotation} \label{subsection:qannot}
\begin{table*}[t!]
%\begin{center}
\small
%\begin{tabularx}{\textwidth}{@{}l C{0.7} C{0.3}@{}}
%\begin{tabularx}{\textwidth}{@{} l X X @{}}
\begin{tabular}{@{}l p{8.4cm}p{4cm}@{}}
\toprule %\hline
\multirow{2}{*}{\makecell{Question Type \\(\# samples)}} & \multicolumn{2}{c}{Example} \\\cmidrule{2-3}%\cline{2-3} 
 &  [\emph{context}] [\emph{source sentence} with span in \textbf{bold}] & Question \\ 
 \midrule %\hline
\multirow{1}{*}{Explanation (443)} & [\dots unraveling of the on-again, off-again UAL buy-out slammed the stock market.][Now, stock prices seem to be in a general \textbf{retreat}.] & Why are the stock prices retreating? \\\\
%& In what may be a revolution in experiencing music, 19-year-old Russian pianist Osip Nikiforov is recording Chopin's Etude Op. 10, No. 1, without capturing any of its sound. \textbf{Instead,} a sensor-equipped piano is recording the ``data'' of his performance, the mechanical movements when keys and foot pedals are pressed. & Why is there a sensor-equipped piano recording data of his performance?\\\hline
\multirow{1}{*}{Elaboration (364)} & [\dots Beth Capper has gone without food 
% to keep up her supply 
\dots][It's not drugs or alcohol or even baby formula that has \textbf{put her in such a bind}.] & What has put her in this bind? \\\\
% & Miami Shores, Fla., tech \textbf{consultant} Rudo Boothe, age 33, attributes his professional success - anyone's professional success, actually - to having learned to read and perform basic math at age 4. & For what company?\\
% & The Agriculture Department says Americans seem to be eating a bit more each year but are \textbf{choosier} about what's on the menu. & what are they choosing?\\
 %& One of Ronald Reagan's attributes as President was that he rarely gave his blessing to the claptrap that passes for ``consensus'' in various international institutions. In fact, he liberated the U.S. from one of the world's most \textbf{corrupt} organizations -- UNESCO. & How is UNESCO corrupt?\\\hline
\multirow{1}{*}{Background (407)} & [\dots John R. Stevens, \dots, was named senior executive vice president\dots][He \textbf{will continue} to report to Donald Pardus, \dots] & How long has he been reporting to Donald Pardus? \\\\
%& BOGOTA, Colombia - In the parking garage of a small apartment building across the highway from Bogota's El Campin soccer stadium, a young man and his mentor practice bullfighting techniques under the \textbf{light} of an atrium. & Are they practicing at night?\\\hline
\multirow{1}{*}{Definition (114)} & [Oh, that terrible Mr. Ortega.][Just when American liberalism had pulled the \textbf{arms plug} on the Contras \dots] & What is the arms plug? \\\\
% & LOS ANGELES - The booming illegal international wildlife trade forced \textbf{conservationists} to do the unthinkable Tuesday: Brand the golden domes of two of the rarest tortoises on Earth to reduce their black market value by making it easier for authorities to trace them if stolen. & Who were the conservationists?\\
%& People start their own businesses for many reasons. But a chance to fill out sales - tax records is rarely one of them. Red tape is the \textbf{bugaboo} of small business. & what is a bugaboo?\\\hline
\multirow{1}{*}{Instantiation (159)} & [\dots in their office, Rajiv Maheswaran and Yu-Han Chang can catch a glimpse of Staples Center \dots][Whiteboards inside their office are filled with \textbf{algorithms} in shades of red, blue and green.] & what kind of algorithms? \\\\
%& The Bush administration's nomination of Clarence Thomas to a seat on the federal appeals court here received a blow this week when the American Bar Association gave Mr. Thomas only a ``qualified'' rating , rather than ``well qualified.'' People familiar with the Senate Judiciary Committee, which will vote on the nomination, said some \textbf{liberal members} of the panel are likely to question the ABA rating in hearings on the matter. & Which liberal members are likely to question the ABA ratings?\\\hline
\multirow{1}{*}{Forward-looking (31)} & [The federal government would not actually shut down. Agents would still patrol \dots][Mail carriers would \textbf{still deliver mail}.] & Would it arrive on time? \\ \\
% & Bethlehem Steel Corp. has agreed in principle to form a joint venture with the world's second-largest steelmaker, Usinor-Sacilor of France, to modernize a portion of Bethlehem's ailing BethForge division. The venture, which involves adding sophisticated equipment to make cast-iron mill rolls, is part of a two-pronged effort to shore up a division that has posted continuing operating losses for several years. The other element includes consolidating BethForge's press-forge operations. The entire division employs about \textbf{850 workers}.& How will they need to increase or decrease staff?\\\hline
\multirow{1}{*}{Other (32)} & [\dots the entire neighborhood can fall victim.] [At this stage some people just \textbf{``walk away''} from homes\dots
% where the mortgage exceeds current market value.
] & Why is it quoted? \\%\hline 
% & Reed International PLC said that net income for the six months ended Oct. 1 slipped 5\% to \# 89.7 million (\$ 141.9 million), or 16 pence a share, from # 94.8 million (\$149.9 million), or 17.3 pence a share. The British paper, packaging and publishing concern, said profit from continuing lines fell 10\% to \# 118 million from \# 130.6 million. While there were no one-time gains or losses in the latest period, there was a one-time gain of \# 18 million in the 1988 period. And while there was no profit this year from discontinued operations, last year they contributed \# 34 million, before tax. Pretax profit fell 3.7\% to \# 128 million from \# 133 million and was below analysts' expectations of \# 130 million to # 135 million, but shares rose \textbf{6 pence to 388 pence} in early trading yesterday in London. & There is no question assuming the first one has been answered. If it hasn't them the first question again.\\
%& QINGDAO, China - As far as Li Lejun is concerned, there's one easy way to make a July beach vacation even better than expected: Add seaweed. Hundreds upon hundreds of tons of it. Buried up to his thighs in sand, his back covered in what looked like strands of \textbf{chartreuse cotton candy,} the 7-year-old Beijing boy was having the time of his life Sunday at No. & Does seaweed look like cotton candy?\\\hline
%
\bottomrule
%\end{tabularx}
\end{tabular}
%\end{center}
\caption{\label{Qtype_distro} Annotated question type distributions and salient examples of each question type. Context and source sentences are presented where the spans in source sentences are bold. More examples are in the Appendix. 
}
\end{table*}

\iffalse
\begin{table}[t!]
\begin{center}
\small
\begin{tabular}{cc}\hline
Model Input & Dev Acc (\%)\\\hline
source+span & 34.7 \\
context+source+span & 36.7 \\
context+source+span+question & 73.3 \\\hline
\end{tabular}
\end{center}
\caption{\label{Qtype_pred} Best dev accuracy of different model inputs for question type prediction, where `context' denotes previous context, and `source' denotes source sentence.}
\end{table}
\fi 
In the USA, K-12 standards describe what students should understand and be able to do by the end of each grade.\footnote{\scriptsize{\url{http://www.corestandards.org}}} The guidelines state that even in very early grades students should understand how individuals and events evolve and interact in a text. The \emph{hows} and \emph{whys} of the text (i.e., inquisitive questions) come naturally to us \cite{ko2020inquisitive}. % \Kevinin{Debanjan, can you add a citation here? see comment from R1}

%We also notice, curiosity driven questions -- such as asking for background information, elaborating details, and why one action led to another -- are linked to Rhetorical Structure Theory (RST). In RST, relations such as \emph{background}, \emph{elaboration}, and \emph{cause} provide a systematic way to analyze the text and understand the discourse relations among segments of the text. Likewise, the questions generated in this work inquire about the background or causal information and those are close to the rhetorical relations in the text. 
%\Debanjan{The above needs to frame again} 

\newcite{ko2020inquisitive} evaluated the question types over a small set of 120 questions and identified a few question types that appear frequently and address various \emph{how}  and \emph{why} questions.\footnote{The evaluation is not available in the released dataset.} Although they presented a fully data-driven approach without any theoretical underpinnings we notice such curiosity driven questions -- such as asking for background information, elaborating details, and why one action led to another -- are linked to Rhetorical Structure Theory (RST) \cite{mann1988rhetorical}. In RST, relations such as \emph{background}, \emph{elaboration}, and \emph{cause} provide a systematic way to analyze the text and understand the discourse relations among segments of the text. Likewise, the questions generated in this work inquire about the background or causal information and those are close to the rhetorical relations in the text. For our annotation, we use the same set of question types as \newcite{ko2020inquisitive}, which are described below: 
%. Below is a short discussion of the question types.  
%
%.\footnote{The evaluation is not available in the released dataset. \Kevinin{We should ask them for that data}} They identified a few question types that appear often in their dataset\Kevinin{what were they?}\Lingyuin{We use same question types as theirs, i.e. Why questions (explanation), Elaboration, Definition, Background information, Instantiation, Forward looking, and Others}. 
%
%Our RST based question type annotation is similar to their ontology. Below we discuss our question types briefly:
%
%While the question types listed in \citep{ko2020inquisitive} are insightful to represent the information that readers are seeking, only 120 questions from 37 sentences are analyzed in that paper without being released, which makes it impossible to use. Therefore, we conduct annotations on question types inspired by their categories, which are listed as below:
%
%\Lingyu{I tried to paraphrase some sentences, but these descriptions are very similar to those in INQUISITIVE. However, I think it important to describe them here.}
\begin{itemizesquish}
    \item \textbf{Explanation}: 
    %(or causal questions): 
    Questions signaled by the interrogative ``why'' as well as its paraphrases such as ``what is the reason''. These questions are often asked to explain why something happened or identify its cause (``why did he choose to speak to the press?'').

    \item \textbf{Elaboration}: Questions that seek more details about concepts, entities, relations, or events expressed in the text, e.g., ``what are some details about this performance?''

     \item \textbf{Background}: Questions that seek more information about the context of the story or seek clarification about something %specific events that are 
     described in the text %without much context 
     (``how much loan was guaranteed?''). 
     %These may be clarification questions. 

    \item \textbf{Definition}: Questions that ask for the meaning of a specific term 
    % keyword that is domain specific. 
    %e.g., 
    (``what does hubris mean?"). 
    
    \item \textbf{Instantiation}: Questions that ask about a specific instance (e.g., ``what is the name of the newspaper?'') or a set of instances (e.g., ``who are these other cable partners?''). %\Kevinin{Changed ``entity'' to ``instance'' as I don't remember there being a requirement that this type involves entities.. I annotated it more generally as any instance of something, whether or not it was an entity.}

    \item \textbf{Forward-looking}: Questions that ask about future events, e.g., ``would it arrive on time?''

    \item \textbf{Other}: Other types of questions, e.g., inference questions (``how many women were found?'') that ask to deduce information from the source, or that ask something irrelevant (``Does seaweed look like cotton candy?'') 
    %\Kevinin{give inference question example here from our annotated data} 
    %For example, given sales reports from two years a question may be ``How much \emph{more} sales were performed this year?'' 
    %\Debanjan{add a gold question.}
    %As another example, given a mention of something without any information about when it occurred, 
    %Given a joint project hailed without mention time, 
    %a question may be ``what year was this?''. 
    %\Kevinin{I edited the text above to try to trim away unnecessary information}
    % Sheraton Corp . and Pan American World Airways announced that they and two Soviet partners will construct two " world - class " hotels within a mile of Red Square in Moscow .	U . S . and Soviet officials hailed the joint project as a new indication of the further thaw in U . S . - Soviet relations . " This is an outstanding example of how the East and the West can work together for their mutual benefit and progress , " said Soviet Ambassador Yuri Dubinin , who hosted a signing ceremony for the venture ' s partners at the Soviet embassy here .
    %\Lingyu{The first question is not a gold question (I didn't find it)? Should we delete the first or give two examples here? Also, the corresponding context and source is commented above this comment.}\Kevinin{I commented out the non-gold inference question. Lingyu, can you find an actual question that we annotated as an inference question to include?}
    
\end{itemizesquish}
%\Debanjan{check with ko et al. also see any other type can be included in the other.}
%\Kevin{I seem to remember seeing some generic high-level questions like ``Of what year?'' but I agree that it would be to good to give an example of something annotated as ``Other'' that is not an inference question.}
%\Lingyu{Ko et al. examples are: `Are they really?', they also listed questions that are subjective, e.g. `Why is doing business in another country instead of America such a sought-after goal?'.}

Three expert annotators who are experienced at annotation tasks initially annotated 50 questions with the types above. 
Pairwise $\kappa$'s between annotators were 0.570, 0.572, and 0.872 %, respectively 
(moderate and substantial agreement). 
%, respectively) 
%\Kevinin{to do: say that we used a majority vote for those first 50}
The annotators exchanged notes and decided on final annotation guidelines. In the next round, each annotator independently annotated 500 random questions from the training partition of \inquisitive, thus producing a total set of 1,550 annotated questions. We used majority vote for the first 50 questions. Table~\ref{Qtype_distro} presents the question type distribution with salient examples. 

\begin{table*}[t!]
\centering
\small
\begin{tabular}{lllllll}\toprule
Explanation & Elaboration & Background & Definition & Instantiation & Forward-looking & Other \\\midrule
why (396) & what (164) & what (108) & what (95) & what (62) & what (9) & why (5) \\
what (28) & how (135) & how (91) & does (5) & which (50) & how (8) & does (5) \\
is (5) & is (11) & is (40) & how (3) & who (36) & will (3) & is (4) \\
how (4) & where (6) & who (34) & who (2) & in (3) & would (2) & what (3) \\
if (3) & in (5) & where (18) & definition (2) & at (2) & did (2) & of (2) \\\bottomrule
\end{tabular}
\caption{\label{ngram_annot} 
%\Kevinin{question on this: it's interesting that the most common bigrams for Background don't include more that start with ``what''.. Lingyu, can you send out the full list of leading bigrams for Background? maybe there are lot of things like ``what [content-word]'', e.g., ``what company'' etc.} \Lingyuin{what was/is/are/did/do all have larger counts, but not so high as ``who is''.}
Most common leading unigrams
% and bigrams 
%\Kevinin{isn't it just ``most common leading unigrams and bigrams'' rather than WH question words? it sounds like we are restricting our attention to WH question words..} 
in annotated questions (lowercased) for each type (counts in parentheses). 
%\Kevinin{We could move the bigrams to the appendix to save space.}\Lingyuin{done.}
%\Lingyuin{These are just most common leading unigrams/bigrams. I'm not sure whether we should write WH question words to match the text. Also, when there’s a tie, should I prioritize those “when, what, why, how, where, who, which” manually? I saw in literature that they do contain “how” into WH words, but maybe not other words (e.g. in, at).}
%\Kevinin{gotcha thanks! we should just say what we are showing here, which is leading unigrams/bigrams (I changed the text accordingly). If there are ties, we should just break ties arbitrarily or alphabetically, rather than doing so based on importance. Also, I assume these counts are computed after converting all questions to lowercase? We should state that here in the caption.}
%\Lingyuin{Yes, these are all lowercased. The table now is automatically generated by code. Though I shuffle order of ``yes what'' with ``how many'' for ``Others'' as they both appear only once.}
}
\end{table*}
Table~\ref{ngram_annot} shows the most common leading unigrams for each question type in our annotated data.\footnote{See appendix for bigrams.}
Although WH question words such as ``why'', ``when'', ``who'', etc.~have been used to generate a variety of question types before \cite{zhou-etal-2019-question}, they cannot fully express the semantic content of  questions \cite{cao-wang-2021-controllable}. Likewise, we observe there is no one-to-one relationship between WH words and question types. Each type encompasses multiple question words. Some types, like Explanation and Definition, have a single dominant leading unigram, while others have two or three. The word ``what'' is the most common leading unigram for five question types. 
\subsection{Question Type Prediction} \label{subsection:qtypes}
%Controllable generation entails having access to an input $x$, an output $q$, and a \textbf{control code} $c$, where the goal is to model $P_\theta(q\mid x,c)$   \cite{kikuchi2016controlling,hu2017toward,keskar2019ctrl}. %\Kevinin{I think we should cite earlier papers from 2017 (or earlier) that did controlled generation} 

%In this work, 
We aim to generate a question that follows a particular question type as control code. However, to do so, we must first determine the question types in the entire \inquisitive dataset. To this end, we finetune RoBERTa-large as a multi-class classifier on the annotated set of 1,550 questions and use the classifier to predict the question types of the remaining questions in \inquisitive. 
%We use RoBERTa-large, a transformer model that performs strongly in standard tasks \cite{liu2019roberta}.
As input, we concatenate the context, source sentence, span, and question, using the  ``[SEP]'' token as delimiter.\footnote{We use a special token ``NO\_CONTEXT'' if the source sentence is the first sentence in the article.} We use 1,400 examples for training and the remaining 150 as the validation set 
(also used for early stopping)\footnote{We keep the distribution of question types in train and dev set roughly the same, and the majority question type (Explanation) is about 29\% of the total data.}, on which we reach an accuracy of 73.3\%. %\Kevinin{to do: say how we used the validation set}

\section{Methods} \label{section:method}
In this section, we present our computational approaches for question generation. The input $x$ is a sequence of tokens $x = \langle x_1,...,x_n\rangle$, which may consist of one or more sentences. 
%\Lingyuin{should we point out that it includes one or more sentences?}\Kevinin{Sure.. I also rewrote this paragraph a bit to make it simpler.} 
% \Kevinin{probably better to write this section in the context of the \inquisitive dataset, i.e., say there is a sequence of words rather than actually refer to it as a sentence, as when the context is present in addition to the source it is not always a single sentence} 
%$x$ that consists of sequence of tokens $x = \langle x_1,...,x_n\rangle$ and corresponding 
The output is a question $q$ that consists of sequence of tokens, i.e., $q = \langle q_1,...,q_m\rangle$.   
%\Kevinin{I changed set notation ($\{\}$) to $\langle \rangle$ because the latter is better for sequences (sets are not ordered).}
Using the standard autoregressive sequence-to-sequence architecture \cite{sutskever2014sequence}  
we model $P_\theta(q\mid x)$ as follows:
\begin{equation} \label{eq:1}
    P_\theta(q\mid x) = \prod_{i}P_\theta(q_i\mid q_1,\dots,q_{i-1},x) 
\end{equation}
%\Lingyu{Equation 1 and 2 changed.}
\noindent We use the pretrained BART model  \cite{lewis-etal-2020-bart}, a transformer \citep{DBLP:conf/nips/VaswaniSPUJGKP17}  composed of a bidirectional encoder and an autoregressive decoder. 
%Below is a short description of the models we used. 
%\Kevinin{We can comment out the following sentence to save space (it's not strictly necessary):} 
%We will use the running example from Table~\ref{source_exp} throughout this section for better understanding. 
%
%\Debanjan{decide on the sections. i.e., how to split the css \dots}
%\paragraph{Baseline models.}
In our simplest setup (called \cs), we concatenate the source sentence and the context. %We refer to this setting as \cs.  
%(context-source, denoted as  \cs in Table~\ref{source_exp} and henceforth) 
The next setting also concatenates the span; we refer to it as \css. 
%(context-source-span, denoted as \css in Table~\ref{source_exp} and henceforth) for question generation. 
Each element (e.g., context, span) is separated with the special token ``[SEP]''.
%\footnote{This is similar to \citet{ko2020inquisitive}; however, their best model used GPT-2 for generation.}

\subsection{Controlled Generation}
% \Kevin{conditional generation is a broad term that could be used to refer to standard seq2seq modeling (since it's a conditional model after all), so I changed it to ``controlled generation'' instead.}
Our next set of models use the question types as control codes to guide question generation. 
%govern the style of question-type specific generation. 
%\Kevinin{not sure we should be crediting CTRL with this formulation, as I think there are examples of it that predate the CTRL paper.. I think it's better to discuss the history of controlled generation in the related work section (my diverse generation paper (\url{https://arxiv.org/pdf/1909.13434.pdf}) cites some 2017 papers) and just say here what we did.}
%As stated in the previous section, 
Controlled generation models   \cite{kikuchi-etal-2016-controlling,hu2017toward,ficler-goldberg-2017-controlling,tsai2021style} condition on a control code $c$ in addition to the input $x$ to model the distribution of $P_\theta(q\mid x,c)$. % where $q$ is the generated question. 
Similar to Eq.~(\ref{eq:1}), we can write,
\begin{equation} \label{eq:2}
    P_\theta(q\mid x,c) = \prod_i P_{\theta}(q_i\mid q_1,\dots,q_{i-1},x,c) 
\end{equation}
%\Kevinin{we need to say how we include $c$ in the model. i.e., do we include it as the initial symbol in the output sequence? also, we need to say how it's represented (i.e., using special symbols in the vocab that are not used for anything else or using natural language versions of the types, or something else)}
Text generation conditioned on such control codes, such as sentiment control of movie reviews, style for chatbots, diverse story continuations, etc., have been used effectively in recent research \cite{tu-etal-2019-generating,krause-etal-2021-gedi-generative,roller2020recipes}. We use the same idea for question generation by conditioning on the question type $c$ as identified in Section \ref{subsection:qtypes}. We simply concatenate the question type as an additional token and finetune BART. Using the example from Table~\ref{source_exp}, the input to BART with the question type Explanation would be:\\
%\begin{itemize}
 %   \item 
    The plan places \dots 2 billion [SEP] Santa Fe \dots transaction [SEP] review [SEP] Explanation
%\end{itemize}

\paragraph{Inference.} We specify the question type to generate specific questions. Top-$k$ sampling with $k=5$ is used to generate questions, where the questions are constrained to be from 5 to 30 tokens, with a length penalty 2.0 \cite{ott2019fairseq}. %\Kevinin{to do: add info about fairseq length penalty} \Lingyuin{Done.} 
The length penalty is an exponential penalty on the length, where a penalty $>1$ favors longer generations.
%\footnote{We use the length penalty as defined in the \texttt{fairseq} toolkit \cite{ott2019fairseq}. 
%\Kevinin{I just added this. Lingyu, is this accurate?}\Lingyuin{Yes. By exponential it means that the penalty is used as ``** penalty'', and I pick this exponential saying from huggingface document (and also checks fairseq implementations.)}\Kevinin{Ah, did you use huggingface then? if so, we should just say that we used the length penalty as defined in the huggingface toolkit.}\Lingyuin{No, I use fairseq for all models except for gpt2. I just sometimes read huggingface document for understanding.}

For each test instance, we generate a question for all question types except ``Other''.\footnote{We made this choice because  ``Other''  includes many subtypes, e.g., inference questions and comparisons, giving us only a few examples per type. We leave this to future work.} Table~\ref{source_exp} shows examples of generated questions.

\subsection{Automatic Question Type Selection}
\label{subsection:selection}
As stated in the Introduction section, besides being able to generate a variety of questions based on a single span, another motivation of this work is to identify a single high-quality question or to rank the list of the questions. In case of controlled generation, one challenge is determining what control code to use at inference time when a single output is desired. We explore two ways to choose a single question from the six generated for each input. 

\paragraph{Informative vs.~inquisitive question classifier.} We  consider using a binary question classifier (RoBERTa-large with default parameters) to classify whether a question is from \inquisitive or SQuAD. We view SQuAD questions as more ``informative'' than inquisitive so we hope for this classifier to capture what it means for a question to be inquisitive. We train on the training questions in \inquisitive and an equal number of questions drawn from SQuAD.\footnote{We also attempted to include the source sentences. However, given the differences between the two datasets (WSJ/AP for \inquisitive vs.~Wikipedia for SQuAD), this caused the classifier to focus more on the source sentences than the questions.} 
%The training data is balanced between the two questions set. %: \inquisitive and SQuAD. 
%We subsample questions from SQuAD since it is larger.
At inference time, given one generated question for each type, we choose the one that maximizes 
%rank them using 
the classifier's probability of being inquisitive. 
%Next, the classifier is run against all the questions that are generated based on the specific control code and we select the question with the highest probability. 
Our hypothesis is that an inquisitive/informative classifier 
%a classifier that is trained to identify inquisitive question against informative question then it will be able to 
can serve as a scoring function for selecting the best  candidate from a set of inquisitive questions. For the example in Table~\ref{source_exp} the classifier chose the Definition question ``what is that?'' with the highest inquisitiveness probability. Below we refer to this method as \cssnqt, where the ``s'' indicates that the SQuAD dataset is used.

\paragraph{Pairwise ranking classifier with expert annotations.} 

In this setup, we collect a small set of question ranking annotations and train a pairwise ranking classifier \cite{liu2009learning} to select the best question. First, we randomly select 300 instances from the 1,885-instance test set from \inquisitive. Next, two expert annotators (each with extensive annotation experience) independently ranked each of the six generated questions per instance. 
%each of the six generated questions.
% from these selected sets. 
%Given the context, source sentence, and span, 
The annotators' task was to rank the questions according to their inquisitiveness and relevance to the context, source, and span. 
%overall quality as inquisitive questions. 
%judge the relevancy of the questions, e.g., whether a generated background question or an explanation question is the better choice here? 
%\Kevinin{I changed the previous sentence and commented out the ``i.e.'' part because it seemed a bit distracting. I also felt that the annotation task was more about overall quality as inquisitive questions rather than about relevancy specifically, but please let me know if it was specifically about relevancy as in that case we should change it back.} 
The annotators judged all six questions for each instance and identified at least three questions (rank 1-3) as the best 
%most relevant ones 
where the rest of the questions were deemed to be of lower quality. % judged not relevant. 
%\Kevinin{same concern as before about the term ``relevant''} 
In some cases, the annotators even ranked top-five questions (rank 1-5). Precision@1, 2, 3 ranks are 0.70, 0.88, and 0.95 respectively (i.e., in 70\% cases one annotator's top-1 selection was found in the other annotator's top-3 selection).\footnote{Please refer to Table \ref{table:rankexamples} in the Appendix section for examples.}

%Using the ranked questions, 
We then approximate the learning-to-rank problem \cite{joachims2007learning,liu2009learning} with a classification problem, i.e., by training a binary classifier to determine whether one question is better than another. For a single input, let $Q$, $q_{\mathit{rel}}$, and $q_{\mathit{nrel}}$ represent the total set of generated questions, relevant questions, and irrelevant questions, respectively. 
In our pairwise ranking setup, the training instances are the 
%\Kevinin{I think ``each training instance is represented as a'' should be changed to ``the training instances are the'', because we train on all instances from both (a) and (b), right?} 
combination of (a) a question $q_{\mathit{i}}$ from $q_{\mathit{rel}}$ and a question $q_{\mathit{j}}$ from $q_{\mathit{nrel}}$, and 
%(line 2-6 in Algorithm 1) 
(b) two questions $q_{\mathit{i}}$ and $q_{\mathit{j}}$  from $q_{\mathit{rel}}$ if and only if the two questions are separated by $\ge$2 ranks. % (line 8-16 in Algorithm 1). 
Algorithm 1 in the appendix details the procedure. 

In addition to the two questions $q_{\mathit{i}}$ and $q_{\mathit{j}}$, we also use the source sentence as another input. During training, for each instance from (a) and (b) above, we create two training examples of the form \emph{source} + ``[SEP]'' + $q_{\mathit{i}}$  + ``[SEP]'' + $q_{\mathit{j}}$ and \emph{source} + ``[SEP]'' + $q_{\mathit{j}}$  + ``[SEP]'' + $q_{\mathit{i}}$. 
%\Kevinin{Do we include any other fields, like the context/source/span?} 
If %$q_{\mathit{i}}$ 
the first question in the sequence has a better rank we label the instance as positive, otherwise negative. This way we have 2,867 examples; we use 2,581 for training and the rest for validation. We finetune a RoBERTa-large model as a binary classifier 
%\Lingyuin{Didn't specify size here as the size is mentioned in the last sentence from the same paragraph.}\Kevinin{I moved this text to be later in this paragraph so that we only mention RoBERTa once in the paragraph; it seemed to flow better this way} 
%. The RoBERTa large classifier 
with default hyperparameters, attaining a validation accuracy of 76.2\%. 

For each test instance, similar to the training setup, for each generated question pair $q_{\mathit{i}}$, $q_{\mathit{j}}$ we form a pair of examples. Given that we have six question types, we create altogether thirty examples and classify them using the RoBERTa-large classifier. 
%\Kevinin{Clarify what the test set mentioned here is. Is it the original \inquisitive test set minus the 300 annotated instances?}.  
We return the question that is preferred the largest number of times.\footnote{In case of ties, we use the classifier scores as the tie breaker.} 
%Next, from the test set\Kevinin{Clarify what the test set mentioned here is. Is it the original \inquisitive test set minus the 300 annotated instances?}, for each test instance, we create a pair of questions. altogether thirty question pairs from six questions per test instance) 
%and classify them using the RoBERTa classifier. Let question $q_i$ and $q_j$ be two questions for the test instance $t_n$. \Kevinin{Again, I think we can remove the $t_n$ notation and just say ``for a test instance''} 
%Then, we will classify $q_i$  + ``[SEP]'' + $q_j$ as well as $q_j$  + ``[SEP]'' + $q_i$. 
%If the classifier returns one for the first case and zero for the second, $q_i$ is preferred over $q_j$. If it is the opposite, then $q_j$ is preferred. If the classifier returns one for both combinations, we have a tie. \Kevinin{How about if the classifier returns zero for both combinations?} 
%\Lingyuin{Sorry but we directly use the how many times first, then use the actual classifier scores as a tiebreaker.} We measure how many times a question is preferred over the other and for simplicity, we simply select the question that is preferred the most number of times per test instance. In case there is still a tie, we break the tie with a random choice. \Kevinin{Oh I thought we were using the actual classifier scores as a tiebreaker.. I probably misremembered} 
Given the example in Table~\ref{source_exp} this model selects the Explanation question, i.e., ``Why are they reviewing the plan?'' Below we refer to this method as \cssrankqt, where the ``r'' represents the use of the ranker described above.

%\Debanjan{a picture explaining the algorithm for the data selection process for above will be helpful.}
%. In this setup, each instance is either a random combination of  (a) a question from the top-3 ranked questions and a     

\iffalse
\begin{figure}[t]
\centering
    \begin{subfigure}{.45\textwidth}
        \centering
        \includegraphics[width=\linewidth]{figures/diagram_c.png}
        % \caption{Conditional generation.}
    \end{subfigure}
    % \begin{subfigure}{.45\textwidth}
    %     \centering
    %     \includegraphics[width=\linewidth]{figures/diagram.png}
    %     \caption{External rescorer.}
    % \end{subfigure}
\caption{\label{rescorer} Diagram for the conditional generation.}
\end{figure}
\fi 
%\subsection{Conditional Generation}

% Naturally, when we seek to generate deeper questions, such as those asking for background information or elaborations, we would want to provide the model with salient features in the text, e.g. entities, aside from the source sentences.
\iffalse 
For conditional generation, we use the ``[SEP]" token as the separator to concatenate all elements including the control codes:

\begin{equation}
    p_{lm} (y|x) \rightarrow  p_{lm} (y|x,c)
\end{equation}
\fi

% We extract the entities and entity types with Stanza \citep{qi-etal-2020-stanza}, and predict the relations between every two entities with OpenNRE \citep{han-etal-2019-opennre}. Then we apply these extra information as control codes, and concatenate them with the source sentences with a separator ``[SEP]'' before input to the model. Examples are shown in Table~\ref{source_exp}.

\section{Experiments} %Results and Discussions} \label{section:result}
\begin{table*}[t!]
\centering
\small
\setlength{\tabcolsep}{4pt}
\begin{tabular}{cccccccccc}\toprule
Model & \%BLEU & \%METEOR & \%ROUGE-L & \%F\textsubscript{BERT} & GPT2 ppl & Entropy & Train-2 & Article-2 & Span   \\ \midrule
%\citet{ko2020inquisitive} & - &- &- & -& -& -& 0.627 & 0.147 & 0.278 \\
\human & -  & -  & -  & - & 272 & 0.777 & 0.467 & 0.126 & 0.354 \\\midrule
%context-source (\cs) 
\cs 
& 4.3 & 11.8 & 27.4 & 39.6 & 119 & 0.699 & 0.518 & 0.186 & 0.184\\% &  7.924 (7)   \\
%context-source-span (\css) 
\css 
& 8.5 & 17.5 & 36.1 & 47.6 & 148 & 0.726 & 0.505 & 0.182 & 0.452 \\% & 8.457 (8) \\
%context-source-span-nqtype (\cssnqt) 
\cssnqt 
& 5.7 & 13.6 & 30.9 & 41.6 & 219 & 0.823 & 0.530 & 0.090 & 0.346 \\% & 6.838 (6)  \\
%context-source-span-rank (\cssrankqt) 
\cssrankqt 
& 8.6 & 18.3 & 35.3 & 47.4 & 89 & 0.612 & 0.473 & 0.195 & 0.542\\% & {8.486} (8)  \\ 
%context-source-span-rqtype (\cssrqt) 
\cssrqt 
& 9.7 & 19.5 & 39.1 & 50.1 & 154 & 0.751 & 0.488 & 0.149 & 0.475 \\% & 7.479 (8)  \\
\bottomrule
\end{tabular}
\caption{\label{traditional_metrics} Automatic metrics on our test set for our models as well as the reference questions (\human). 
}
\end{table*}

For all models, 
%\cs, \css, \cssnqt, \cssrankqt, and \cssrqt, 
we use BART-large with the same settings. For training, we use the Adam optimizer \citep{DBLP:journals/corr/KingmaB14} with learning rate 3e-5, weight decay 0.01, clip norm 0.1, dropout 0.1, 15 epochs in total, warm-up updates 500, use cross entropy loss with label smoothing ($\alpha = 0.1$), and set the maximum number of tokens per batch to 1024. More details of the experimental setup are given in the Appendix (Section \ref{subsection:expparameters}). 

We evaluate the following five settings:
\begin{itemizesquish}
    \item \cs: uncontrolled generation using the context and source sentence as input
    \item \css: uncontrolled generation using the context, source sentence, and span as input
    \item \cssnqt: type-controlled generation with type selection via informative vs.~inquisitive classifier
    \item \cssrankqt: type-controlled generation with type selection via pairwise ranking classifier
    \item \cssrqt: type-controlled generation with question type of reference question 
\end{itemizesquish}
\noindent Since the \qtfamily methods use question types, in order to compare those methods to others, we need a way to automatically select a single generated question. 
For \cssrqt, we run our question type classifier on a human-written reference question and use the predicted type. Thus, \cssrqt is an oracle method (hence the mnemonic ``o'' in its name) since it assumes access to a reference question. 
For \cssnqt and \cssrankqt we use the classifiers described in Section~\ref{subsection:selection}. All  \qtfamily methods use the context, source sentence, and span as input, like \css.

\subsection{Automatic Evaluation} \label{subsection:autoeval}
%Inquisitive question generation is an open-ended generation task. Since there are many possible questions to ask about a given span, 
Since inquisitive question generation is an open-ended task, 
a high-quality generated question may not overlap with the gold question. However, automatic metrics that measure the overlap between generations and gold questions could still be useful diagnostics for characterizing models. 
%throw light on intrinsic connections. 

Table~\ref{traditional_metrics} presents several automatic metrics: 
%scores as measured via standard metrics such as 
BLEU \citep{papineni-etal-2002-bleu}, 
METEOR \citep{denkowski-lavie-2014-meteor}, 
ROUGE-L \citep{lin-2004-rouge}, 
BERTScore \citep{DBLP:conf/iclr/ZhangKWWA20},
perplexity under GPT2-XL 
%\Kevinin{clarify here which size of GPT2 we used}\Lingyuin{Finish updating with GPT2-xl.} 
\citep{radford2019language}, and the entropy (averaged over questions) of the RoBERTa-large question type classifier applied to the generated question.\footnote{We reported the average scores of 5 runs with different random seeds.} 
Although \inquisitive contains a test set of 1,885 instances (see Section \ref{section:data}), we used only 1,585 instances as our test set because we chose the remaining (random) 300 instances to build our 
pairwise ranking classifier 
%for automatic question type selection 
(Section \ref{subsection:selection}).
%As stated earlier, although the test partition from \inquisitive has 1,885 instances (see Section \ref{section:data}), here we report the results for 1,585 instances omitting the 300 instances used to build the pairwise ranker classifier.

%Aside from \cssnqt, the other models have higher BLEU scores than \cs, and \cssrankqt has the highest. %attains the highest score. 
% We observe, beside \cssnqt, rest of the models report comparable BLEU scores where the \cssrankqt model attains the highest score. 
For BLEU, METEOR, ROUGE-L, and BERTScore, the oracle model \cssrqt achieves the highest scores, presumably because this model generates questions that are similar to the reference types. We notice, \cssrankqt, and \css have similar scores, with \cssrankqt being slightly ahead for BLEU and METEOR. In the case of \cssnqt, the low scores across metrics can be attributed to the fact that the inquisitive vs.~informative classifier prefers question types that are unique to the \inquisitive dataset, such as Definition and Instantiation questions. These types are not appropriate for all spans and in many cases are quite different from the reference questions. 
%It is not always appropriate to generate a definition or instantiation question depending on the span. 

%We also consider computing perplexity (using GPT2-XL) and entropy of the question type classifier for the models and notice the 

We also find that \cssrankqt has the lowest GPT2 perplexity, indicating that the ranker is favoring highly probable questions according to a  general-purpose language model. A lower perplexity is likely indicative of greater fluency, a point we will return to in our human evaluations. 
%likely under the underlying probability distribution is good at predicting the question type 
%\Lingyuin{Could we say that GPT-2 is related to fluency?}\Kevinin{yea we can roughly say this.. i edited it a little}
%\Kevinin{a frequent edit I made in this section: a model can't ``report'' a result; we are the ones who are doing the reporting}
Likewise, the lowest entropy of \cssrankqt implies that its questions can be classified with high confidence by our question type classifier. 
%are easier to classify to a specific question type, e.g., Explanation. 
%If we report question type classifier entropy loss, \cssrankqt has the lowest score, which implies that people tend to select questions that are easier to be classified into a specific question type, e.g. Explanation questions. 
In contrast, \cssnqt
%, which often selected Definition/Instantiation questions, 
shows higher entropy, i.e., its questions are more difficult to classify. The entropy of the human-generated questions is higher than nearly all of our models, indicating that the human questions are also harder to classify than model outputs. 
The last three columns of Table~\ref{traditional_metrics} show the metrics designed by \newcite{ko2020inquisitive}, namely \emph{Train-n}, \emph{Article-n}, and \emph{Span}. These metrics measure the extent of copying from the source materials into the generated questions, i.e., \% of $n$-grams in the generated questions that appear in the training questions (\emph{Train-n}) and the context/source sentence (\emph{Article-n}). 
For brevity, we only report \emph{Train-2} and \emph{Article-2}. 
\emph{Span} measures the \% of words in the annotated span present in the generated questions.  

%\Lingyuin{I think Article-n measures the relativeness of the question to the related article? So we expect higher Article-n and lower Train-n in some way. I'm not sure whether the description here would make people think we want both be lower.}\Kevinin{I'm not convinced that we want any of these metrics to be high or low.. perhaps one could argue that we would want to match the human values, but in any event I don't think higher or lower is ``better''.. these are just ways to characterize models rather than metrics to be optimized}.

%\emph{Train-n} and \emph{Article-n} 

%measure the \% of $n$-grams in the generated questions appeared in the questions in the training set  and  
%Likewise, \emph{Article-n}  measures the \% of $n$-grams from the source sentence or the context  

%articles \Kevinin{What exactly is an ``article'' here? Is it the concatenation of the context and the source for a single instance? Or just the context? Is it the concatenation of all training data contexts (or context-source concatenations) or only for a single instance? Please add more text to clarify.} that are present in the questions. Finally, \emph{Span} measures the \% of words in the annotated spans \Kevinin{instance level or training-set level?} that are also in the questions. 

Among our models, \cssrankqt attains the lowest value of the \emph{Train-2} metric, which is also closest to the \human value. 
%report lower \emph{Train-n} values than \citet{ko2020inquisitive}.\footnote{We recomputed \human scores here since our test set is smaller than the \inquisitive test set.}
%\Lingyuin{In fact, I'm not sure whether comparing our scores with Ko et al would work. They report on full test set and we report on part of the test set. Also, when I try to compute Human scores with full test set, the results are different from their computations (more close to the scores we reported here). Their human scores listed in paper are: Train-2 0.520, Article-2 0.115, Span 0.219. I guess this might due to different tokenization methods, am trying to match their scores now.
%}\Kevinin{sounds good}\Lingyuin{Tried re, spacy, and nltk. nltk is the closest but still different. Maybe that's all I could do now.}
%\Kevinin{Have they released any code? If not, we should email them to ask them how they exactly computed those metrics. I agree that we can't compare directly to Ko et al if their numbers were computed using a different tokenization or otherwise a different implementation}
% , whereas the \cssrankqt consistently attains lowest scores
%Regarding \emph{Article-2} metric, we observe mixed results, i.e., except 
Aside from \cssnqt, the other models have higher \emph{Article-2} than \human, meaning that the generated questions have a higher \% of $n$-grams that appear in the source sentence or the context.
%Here, the \css model obtains the highest score. 
\cssrankqt has the highest value for the \emph{Span} metric, indicating that the ranker prefers questions that use words from the span. 
%\cssrankqt and \css have the highest values. 
\cssrqt is second highest and \cs, which does not use the annotated span, has the lowest value. 
%, which is expected because this model did not use the annotated span for question generation. 
%Across all metrics, \cssrankqt stood out - either the question generated are the most novel (e.g., \emph{Train-2} score) or used the highest \% of n-grams from the source text for generation. 
%\Kevinin{There was a sentence here that was not true for the Article-2 metric though so I commented it out}

In the Appendix, we also report an automatic evaluation of controllability, finding that certain question types (Explanation, Definition, and Instantiation) can be generated with high precision, while others (Elaboration, Background, and Forward-looking) are more easily confused.

\subsection{Human Evaluation} \label{subsection:humeval}

%In the previous section, our results demonstrate that our models generate novel questions that differ from those in the training set. Likewise, they generate questions that overlap with the source texts and spans. 
In this section, we report the results from a human evaluation we have conducted to assess a variety of subjective aspects of the generated questions, namely the \emph{syntax}, \emph{semantics}, \emph{relevancy}, and the degree of \emph{inquisitiveness}. 
%of the generated questions. 

\iffalse
%the following table is based on ``max'' vote
\begin{table}[t!]
\centering
\small
\setlength{\tabcolsep}{4pt}
\begin{tabular}{lllll}\toprule
Model & Syn. & Sem. & Rel. & Inq. \\\midrule
\cs & 4.50 & 4.22 & 4.34 & 3.87 \\
\css & 4.49 & 4.30 & \textbf{4.45} & 4.02 \\
\cssnqt & 4.12 & 3.50 & 3.57 & 3.21 \\
\cssrankqt & \textbf{4.74} & \textbf{4.58} & \textbf{4.46} & \textbf{4.16} \\
\cssrqt & \textbf{4.51} & {4.32} & 4.34 & \textbf{4.05} \\

\midrule

Human & {4.52} & {4.58} & 4.46 & {4.23} \\
\bottomrule
\end{tabular}
\caption{\label{human_eval} Results from the MTurk judgments. Syntax, Semantics, Relevancy, and Inquisitiveness are abbreviated as Syn., Sem., Rel., and Inq. The \human row shows judgments for reference questions from the \inquisitive dataset. 
%, respectively. 
%High scores are \textbf{bold}. \Kevin{What does ``high score'' mean here? It seems a little arbitrary since 4.05 is bold but 4.02 is not. I think it would be best to compute statistical significance of a few key comparisons to give the reader a sense of which differences are statistically significant.} 
}
\end{table}
\fi
% the following table is based on avg 
\begin{table}[t!]
\centering
\small
\setlength{\tabcolsep}{5pt}
\begin{tabular}{lcccc}\toprule
Model & Syntax & Semantics & Relevancy & Inquisitive \\\midrule
\cs &	4.30	&	4.11	&	4.16	&	3.71	\\
\css &	4.30	&	4.17	&	4.32	&	3.75	\\
\cssnqt &	4.02	&	3.50	&	3.51	&	3.14	\\
\cssrankqt &	4.59	&	4.37	&	4.27	&	3.92	\\
\cssrqt	& 4.33	&	4.10	&	4.09	&	3.78	\\

\midrule

\human &	4.36	&	4.41	&	4.33	&	3.98	\\
\bottomrule
\end{tabular}
\caption{\label{human_eval} Results of human evaluation. The  \human row shows judgments for reference questions from the \inquisitive dataset.} 

%\Kevin{What does ``high score'' mean here? It seems a little arbitrary since 4.05 is bold but 4.02 is not. I think it would be best to compute statistical significance of a few key comparisons to give the reader a sense of which differences are statistically significant.} }
%\Debanjan{changed to avg value - will edit the text later.}
\end{table}

We collected annotations using the crowdsourcing platform Amazon Mechanical Turk (MTurk). We randomly selected 500 test instances. For each, we asked three annotators the following four questions to measure quality along four aspects:   
%Note, instead of using the keywords (\emph{syntax}, \emph{semantics}, \emph{relevancy}, and \emph{inquisitiveness}) we phrased the questions to become more accessible to the Turkers.
\begin{enumeratesquish}
    \item Does the question seem syntactically correct?
    \item Does the question make sense (semantically)?
    \item Does the question seem relevant to the source? 
    \item Does the question show inquisitiveness to learn more about the topic?
\end{enumeratesquish}
The annotators were given the following three options to choose from: \emph{yes}, \emph{somewhat}, and \emph{no}. Each human intelligence task (HIT) 
contained five instances to judge and we paid \$2 per HIT.   
%For 2, it seems a trickier task and we might need to use expert annotators for that?
%Besides, we might need to repeat the above when generated by a different model(s), such as other conditional language models or apply a rescorer(s) to further improvements in generation or using gpt-3 (few shot examples fed).
%We collected judgments for the same set of 500 test instances for the models we reported in Table~\ref{traditional_metrics} and Table~\ref{Ko_metrics}. \Kevinin{I thought Tables 3 and 4 reported results for the 1,588 test instances excluding the 300 used to train the ranker.. the previous sentence makes it sound like Tables 3-4 reported results on a set of 500 test instances?} \Lingyuin{Yes, Tables 3 and 4 reported results for the 1,588 test instances excluding the 300 used to train the ranker.} 
\begin{table}[t!]
\centering
\small
\begin{tabular}{l|p{0.8\linewidth}}\toprule
1 & is it the aha?\\
2 & how much has inflation?\\
3 & nativity happens for buddha?\\
4 & When he decide?\\
%5 & Why is there ability uncertain?\\
5 & how much has inflation? \\
%6 & Why did they order the book seized?\\
\bottomrule
\end{tabular}
\caption{\label{table:wronggold} Examples of gold questions from \inquisitive dataset that are judged as ungrammatical by the Turkers.}
\end{table}

\iffalse
\begin{table}[t!]
\centering
\small
\begin{tabular}{l|p{0.8\linewidth}}\toprule
1 & why did the nine months matter?\\
2 & Why is the VP of Public Affairs for the American Wind Energy Association?\\
3 & How do they prevent the spread of strategic weapons?\\
4 & Why is he not racing on oval tracks before?\\
\bottomrule
\end{tabular}
\caption{\label{table:wrongall} Examples of model generations that are ungrammatical.}
\end{table}
\fi 
\begin{table*}[th]
\begin{center}
\small
\setlength{\tabcolsep}{4pt}
%\begin{tabularx}{\textwidth}{@{}C{0.5} l L %C{0.7} 
%llll@{}}
\begin{tabular}{@{} p{3.5cm} c p{7.5cm} cccc@{}}
\toprule
% \begin{tabular}{p{3cm}p{1cm}p{3cm}p{1cm}p{1cm}p{1cm}p{1cm}}
% \hline
Source  & Model & Question  & Syn. & Sem. & Rel. & Inq.            \\ \midrule
\multirow{6}{=}{\dots The State Security court said it was ordering the seizure of Kemal's book, \dots, because it \textbf{provokes ``hatred and enmity''} on the basis of differences \dots} & \cs & What is the reason for the seizure? & 4.3 & 3.7 & 4.3 & 4.3         \\ %\cline{2-7} 
& \css  & How does the author’s book provoke hatred and enmity? & 5.0          & 5.0      &   5.0        & 5.0        \\ %\cline{2-7} 
 & \cssnqt  & what is hatred? & 4.3         & 4.3        & 2.3          & 1.7 \\ %\cline{2-7} 
& \cssrankqt  & How can a book provoke hatred and enmity?             & 5.0          & 5.0         & 5.0          & 3.0 \\ %\cline{2-7} 
& \cssrqt   &  How did it provocate the book and what did the author write in the book?                                                                     & 5.0          & 2.3          & 3.0          & 2.3          \\ %\cline{2-7} 
& \human  &  How does the book provoke ``hatred and enmity''?  & 4.3          & 5.0 & 4.3          & 3.7          \\ \midrule

\multirow{6}{=}{Olin Corp. said \textbf{third-quarter} net income rose 26 \% on the strength of its chemical business.} & \cs & Why were they on the strength of their chemical business? & {1.0} & 1.0 & {3.0} & 3.0          \\ %\cline{2-7} 
& \css  & what happened to the second quarter? & 3.7          & 3.7       &   4.3          & 3.0          \\ %\cline{2-7} 
 & \cssnqt  & What was the quarter? & 4.3          & 4.3          & 4.3          & {4.3} \\ %\cline{2-7} \
& \cssrankqt  & Why was third-quarter net income up 26 \%?             & 5.0          & 5.0         & 4.3          & 4.3 \\ %\cline{2-7} 
& \cssrqt   & What year did the third-quarter consist of?                                                                  & 5.0          & 5.0          & 4.3          & 5.0          \\ %\cline{2-7} 
& \human  &  What happened to the net income in the first and second quarter?  & 3.7          & {5.0} & 5.0          & 5.0          \\ \midrule

\multirow{6}{=}{\dots most significant change in surgical training since the early 1900s, they are working with \textbf{local medical device companies} to develop new generations of software \dots} & \cs & How are medical device companies working with the University of Minnesota?? & {5.0} & 5.0 & {5.0} & 4.3          \\ %\cline{2-7} 
& \css  & Which local medical device companies? & 2.3          & 3.0      &   4.3          & 3.0          \\ %\cline{2-7} 
 & \cssnqt  & who are the local medical device companies? & 4.3          & 3.7          & 2.3          & {2.3} \\ %\cline{2-7} 
& \cssrankqt  & Why are they working with local medical device companies?             & 5.0          & 5.0         & 5.0          & 5.0 \\ %\cline{2-7} 
& \cssrqt   & Who are the local medical device companies?                                                                     & 5.0          & 3.7          & 4.3          & 5.0          \\ %\cline{2-7} 
& \human  &  Which medical device companies are being worked with?  & 2.3          & {3.7} & 5.0          & 5.0         \\ 
\bottomrule
\end{tabular}
\caption{\label{table:humananalysis} Examples of generated questions from different models. Syn., Sem., Rel., Inq. represent Syntax, Semantics, Relevancy and Inquisitiveness, respectively. For brevity the context is not shown. Spans are \text{bold}.%\Lingyuin{``Human" has different fonts here, so I change it to \human, I could change back if needed. Also, should we remove the lines between questions for the same source sentence?}
}
\end{center}
\end{table*}

Table~\ref{human_eval} presents the average of the human judgments, where the answers \emph{yes}, \emph{somewhat}, and \emph{no} are converted to scores 5, 3, and 1, respectively. 
In all four aspects, we notice several scores are over 4. For the \emph{inquisitiveness} aspect, the \cssrankqt model achieves the highest score among all models. This score is higher than the oracle model (\cssrqt) showing the usefulness of the ranker to generate inquisitive questions. Likewise, \cssrankqt achieves the highest average score for \emph{semantics}, showing that its questions are semantically meaningful almost all the time. We also note that both  \cssrankqt and \css are competitive in \emph{relevancy}. Finally, for \emph{syntax}, each model (aside from \cssnqt) was rated close to 4.5. Although transformers usually produce fluent output  \cite{lin2021pretrained}, \cssrankqt scored higher than the human generated gold questions on \emph{syntax}, which warrants further investigation. 

%Given \cssrankqt and \human are rated with scores of 4.59 and 4.36 (out of 5) in Table~\ref{human_eval}, we realize the small difference between them is probably due to the examples that the Turkers judged as grammatically incorrect. Thus, we 
We manually analyzed all the questions from \cssrankqt and \human where the majority of annotators rated 1 for \emph{syntax}, and we found out there are 14 and 33 such questions, respectively, explaining why \cssrankqt scored higher. This also explains why the perplexity of \human is high in Table~\ref{traditional_metrics}. In Table \ref{table:wronggold} we provide a few examples from the \human set that were deemed grammatically incorrect. Table \ref{table:wrongcssrank_full} in the Appendix contains examples of grammatical errors from all models. 
%\Lingyuin{Should we keep the examples from the model generated? It's in tables/grammar.tex and commented out.}

If we compute a simple average over all the aspects for each model, \cssrankqt scores the highest, slightly better than \human, and the \css model is second. The \css model is trained on the \inquisitive dataset so it has the freedom of generating inquisitive questions. However, what separates \css from \cssrankqt is, for the latter, we have the ability to control the generation with specific question types and also select the \emph{best} question for the same source sentence. 
%\Kevinin{etc.. say that typer is best, slightly better than human, and span is second} 
We also notice that the generations from \cssnqt scored lowest across all four aspects. The \cssnqt model often selects Definition/Instantiation question types that are unsuitable for the source sentence and the span, which is why the annotations score low for this type of question.

Table~\ref{table:humananalysis} shows several examples from our models along with average human ratings for all four aspects. We highlight three salient observations here. First, in general, \cssrankqt has high scores across all  aspects for all examples. Second, the Turkers have treated the aspects independently as we have requested. Even if they rated the \human annotations 2.3 and 3.7 for \emph{syntax} and \emph{semantics} for the last example, they have given high ratings for the other two aspects. Third, interestingly, ``what is hatred?'', a very generic question, scored high on \emph{syntax} and \emph{semantics} (\cssnqt model for the first example) but low on the other two aspects due to its lack of \emph{relevancy} and \emph{inquisitiveness}.    

Finally, we note that for the first example in Table~\ref{table:humananalysis}, the \css and \human questions are extremely similar, but their ratings differ for three out of the four attributes. This example illustrates the variability of human judgments for this task, which suggests that more annotations may be needed to increase confidence in our results.

\section{Related Work} \label{section:related}

In recent years, automatic question generation has attracted many NLP researchers, perhaps due to its versatility, e.g., question generation for conversational AI \cite{bordes2016learning, gao-etal-2019-interconnected}, synthetic examples for QA tasks \cite{alberti-etal-2019-synthetic,DBLP:journals/dad/OlneyGP12,sultan-etal-2020-importance}, clarifications on information-seeking conversation \cite{aliannejadi2019asking}, and knowledge evaluation and educational application areas \cite{mitkov-ha-2003-computer,brown-etal-2005-automatic,Chen2009GeneratingQA,stasaski-etal-2021-automatically}, which is specifically related to our use cases. 

In earlier work, methods such as transforming declarative sentences into questions \cite{heilman-smith-2010-good} or using semantic roles \cite{flor-riordan-2018-semantic} were popular. However, recently sequence-to-sequence architectures \cite{du-etal-2017-learning,fitzgerald-etal-2018-large} and pretrained models \cite{cao-wang-2021-controllable} are more often used. Similar to \citet{ko2020inquisitive}, our work is related to answer-agnostic question generation. We focus on exploiting question type information for generating deeper questions. 
Although related work in the answer-unaware setting exists \cite{nakanishi2019towards}, they mostly focus on identifying question-worthy text for generation \cite{scialom-staiano-2020-ask,wang2019multi} from factual \cite{rajpurkar-etal-2016-squad}, conversational \cite{choi-etal-2018-quac}, or social media platforms \cite{fan-etal-2019-eli5}, different from the WSJ/AP news dataset used in our work. 

We are building on past work on controllable generation, generating text that reflects specific characteristics of control variables. In some earlier work, embedding vectors of the control variables were fed into the model for controlling the output \cite{kikuchi-etal-2016-controlling,fan-etal-2018-controllable,tu-etal-2019-generating}. However, our approach resembles recent efforts where the control variable is concatenated to the main input using some separator \cite{keskar2019ctrl,schiller-etal-2021-aspect}.
%which has been successfully applied to generate diverse story continuations, style transfer for chatbots, controllable summarization, and controlling topic and sentiment polarity, among others  \cite{tu2019generating,kikuchi2016controlling,krause-etal-2021-gedi-generative,wang2017steering}.
%\Kevinin{it would be good to distinguish between citations of methods for controllable generation and citations of applications of controllable generation.. the above sentence is talking about applications, but then citing both the original controllable generation papers as well as application papers; I'm now thinking that applications of controllable generation are out of scope for this paper, since there are way too many. When we cite controllable generation papers, I think we should only cite the earliest methods papers that we essentially follow, as well as papers that involve question generation.} 
Methods such as PPLM are useful for similar guided controllable generations \cite{Dathathri2020Plug}; however, PPLM requires  gradient descent at inference time, while our question type selection approach is highly scalable and efficient. 

We consider controllable question generation based on specific question types, noting that different question templates or ontologies have been studied for question generation. For example, a Wikipedia-driven ontology is used for generation \cite{labutov2015deep}, or contextualized questions are generated for any semantic role \cite{pyatkin-etal-2021-asking}. 
Likewise, \newcite{pascual-etal-2021-plug-play} proposed guided generation focusing on including specific keywords (e.g., ``wh'' words for questions), while we showed in Table~\ref{ngram_annot} that ``wh'' words do not have a 1-to-1 relationship with question types.
%\caption{\label{ngram_annot} 

Our work is closer to that of \citet{cao-wang-2021-controllable}, who proposed a question type ontology (based on cognitive science) inspired by manually constructed templates \cite{DBLP:journals/dad/OlneyGP12}. 
On the contrary, we chose a dataset that focuses on inquisitive questions only and chose our question types accordingly, while they used a dataset with a broader set of questions. 
%different question types and datasets; 
In addition, instead of predicting the text span (``focus'' in \cite{cao-wang-2021-controllable}) we directly use the annotated span in our research. 
%Although our question types are similar, 
Finally, we focused on post-processing the generations to identify the best question (or rank them) related to the source content.

\section{Conclusions and Future Work} \label{section:conclusion}
We proposed a type-controlled framework that generates inquisitive questions given a source sentence, annotated span, and %(occasionally) 
a longer context. We annotated a set of question types related to curiosity driven questions and demonstrated that our framework can generate a variety of questions from a single input. We also developed an effective method (\cssrankqt) to select a single question using a pairwise ranker trained on a small set of ranking annotations. 
%post-processing steps such as a pairwise ranking classifier (\cssrankqt) to select the {best} question. 
Our generations, especially from \cssrankqt, show high novelty. The human evaluation demonstrates that questions generated from \cssrankqt rival human-written questions on all four aspects of quality. 
%the best syntax, semantics, and inquisitiveness scores (aside from the human-written questions). 

%We notice that questions generated in this work inquire about the background or causal information and those are close to the rhetorical relations in the text. 
Future work could include annotating a larger partition of the \inquisitive dataset while exploring finer-grained analysis of question types (e.g., sub-categories of elaboration questions). We are also interested in employing a framework to generate questions and identify the span jointly. 

\iffalse 
TODO: \\
1. significant testing \\ 
Debanjan \\
1. new table with generation examples \\
2. run average rather max vote \\
3.  ``other'' example \\ 
Lingyu:\\
1. run BLEU etc. (done)\\
2. leading n-gram numbers (done)\\
3. gpt-2 large at least (done, use xl)\\
4. appendix for Tables (ongoing)\\
\fi

\section*{Acknowledgments}
The authors would like to thank Yuan Wang and Hillary Molloy for annotations, Mengxuan Zhao and Rob Bobkoskie for the MTurk experiments, and Swapna Somasundaran for  helpful comments and suggestions. 
\section{Ethical Considerations} \label{section:ethics}

We leverage the freely available open access question dataset \inquisitive for annotation and model training. Though we have not exhaustively checked the source dataset manually, given they are sourced from the WSJ partition of the Penn Treebank and Associated Press articles from the TIPSTER corpus, we consider them relatively safe and do not find any objectionable content.  

Training is done using large pretrained models that have been shown to have bias. Although the generated questions do not appear biased, they may hallucinate content, which is a common problem for neural generation models.

Finally, we obtained institutional review board permission to conduct MTurk based data collection.

\bibliography{anthology,custom}%anthology-sm

\begin{thebibliography}{61}
\expandafter\ifx\csname natexlab\endcsname\relax\def\natexlab#1{#1}\fi

\bibitem[{Alberti et~al.(2019)Alberti, Andor, Pitler, Devlin, and
  Collins}]{alberti-etal-2019-synthetic}
Chris Alberti, Daniel Andor, Emily Pitler, Jacob Devlin, and Michael Collins.
  2019.
\newblock \href {https://doi.org/10.18653/v1/P19-1620} {Synthetic {QA} corpora
  generation with roundtrip consistency}.
\newblock In \emph{Proceedings of the 57th Annual Meeting of the Association
  for Computational Linguistics}, pages 6168--6173, Florence, Italy.
  Association for Computational Linguistics.

\bibitem[{Aliannejadi et~al.(2019)Aliannejadi, Zamani, Crestani, and
  Croft}]{aliannejadi2019asking}
Mohammad Aliannejadi, Hamed Zamani, Fabio Crestani, and W.~Bruce Croft. 2019.
\newblock \href {https://doi.org/10.1145/3331184.3331265} {Asking clarifying
  questions in open-domain information-seeking conversations}.
\newblock In \emph{Proceedings of the 42nd International {ACM} {SIGIR}
  Conference on Research and Development in Information Retrieval, {SIGIR}
  2019, Paris, France, July 21-25, 2019}, pages 475--484. {ACM}.

\bibitem[{Bordes et~al.(2017)Bordes, Boureau, and Weston}]{bordes2016learning}
Antoine Bordes, Y{-}Lan Boureau, and Jason Weston. 2017.
\newblock \href {https://openreview.net/forum?id=S1Bb3D5gg} {Learning
  end-to-end goal-oriented dialog}.
\newblock In \emph{5th International Conference on Learning Representations,
  {ICLR} 2017, Toulon, France, April 24-26, 2017, Conference Track
  Proceedings}. OpenReview.net.

\bibitem[{Brown et~al.(2005)Brown, Frishkoff, and
  Eskenazi}]{brown-etal-2005-automatic}
Jonathan Brown, Gwen Frishkoff, and Maxine Eskenazi. 2005.
\newblock \href {https://aclanthology.org/H05-1103} {Automatic question
  generation for vocabulary assessment}.
\newblock In \emph{Proceedings of Human Language Technology Conference and
  Conference on Empirical Methods in Natural Language Processing}, pages
  819--826, Vancouver, British Columbia, Canada. Association for Computational
  Linguistics.

\bibitem[{Cao and Wang(2021)}]{cao-wang-2021-controllable}
Shuyang Cao and Lu~Wang. 2021.
\newblock \href {https://doi.org/10.18653/v1/2021.acl-long.502} {Controllable
  open-ended question generation with a new question type ontology}.
\newblock In \emph{Proceedings of the 59th Annual Meeting of the Association
  for Computational Linguistics and the 11th International Joint Conference on
  Natural Language Processing (Volume 1: Long Papers)}, pages 6424--6439,
  Online. Association for Computational Linguistics.

\bibitem[{Chen et~al.(2018)Chen, Yang, Hauff, and Houben}]{ICWSM18LearningQ}
Guanliang Chen, Jie Yang, Claudia Hauff, and Geert-Jan Houben. 2018.
\newblock {LearningQ}: A large-scale dataset for educational question
  generation.
\newblock In \emph{Proceedings of International Conference on Web and Social
  Media (ICWSM)}.

\bibitem[{Chen et~al.(2009)Chen, Aist, and Mostow}]{Chen2009GeneratingQA}
Wei Chen, Gregory Aist, and Jack Mostow. 2009.
\newblock Generating questions automatically from informational text.

\bibitem[{Cho et~al.(2019)Cho, Seo, and Hajishirzi}]{cho-etal-2019-mixture}
Jaemin Cho, Minjoon Seo, and Hannaneh Hajishirzi. 2019.
\newblock \href {https://doi.org/10.18653/v1/D19-1308} {Mixture content
  selection for diverse sequence generation}.
\newblock In \emph{Proceedings of the 2019 Conference on Empirical Methods in
  Natural Language Processing and the 9th International Joint Conference on
  Natural Language Processing (EMNLP-IJCNLP)}, pages 3121--3131, Hong Kong,
  China. Association for Computational Linguistics.

\bibitem[{Choi et~al.(2018)Choi, He, Iyyer, Yatskar, Yih, Choi, Liang, and
  Zettlemoyer}]{choi-etal-2018-quac}
Eunsol Choi, He~He, Mohit Iyyer, Mark Yatskar, Wen-tau Yih, Yejin Choi, Percy
  Liang, and Luke Zettlemoyer. 2018.
\newblock \href {https://doi.org/10.18653/v1/D18-1241} {{Q}u{AC}: Question
  answering in context}.
\newblock In \emph{Proceedings of the 2018 Conference on Empirical Methods in
  Natural Language Processing}, pages 2174--2184, Brussels, Belgium.
  Association for Computational Linguistics.

\bibitem[{Dathathri et~al.(2020)Dathathri, Madotto, Lan, Hung, Frank, Molino,
  Yosinski, and Liu}]{Dathathri2020Plug}
Sumanth Dathathri, Andrea Madotto, Janice Lan, Jane Hung, Eric Frank, Piero
  Molino, Jason Yosinski, and Rosanne Liu. 2020.
\newblock \href {https://openreview.net/forum?id=H1edEyBKDS} {Plug and play
  language models: {A} simple approach to controlled text generation}.
\newblock In \emph{8th International Conference on Learning Representations,
  {ICLR} 2020, Addis Ababa, Ethiopia, April 26-30, 2020}. OpenReview.net.

\bibitem[{Denkowski and Lavie(2014)}]{denkowski-lavie-2014-meteor}
Michael Denkowski and Alon Lavie. 2014.
\newblock \href {https://doi.org/10.3115/v1/W14-3348} {Meteor universal:
  Language specific translation evaluation for any target language}.
\newblock In \emph{Proceedings of the Ninth Workshop on Statistical Machine
  Translation}, pages 376--380, Baltimore, Maryland, USA. Association for
  Computational Linguistics.

\bibitem[{Dong et~al.(2019)Dong, Yang, Wang, Wei, Liu, Wang, Gao, Zhou, and
  Hon}]{10.5555/3454287.3455457}
Li~Dong, Nan Yang, Wenhui Wang, Furu Wei, Xiaodong Liu, Yu~Wang, Jianfeng Gao,
  Ming Zhou, and Hsiao-Wuen Hon. 2019.
\newblock \href
  {https://proceedings.neurips.cc/paper/2019/file/c20bb2d9a50d5ac1f713f8b34d9aac5a-Paper.pdf}
  {Unified language model pre-training for natural language understanding and
  generation}.
\newblock In \emph{Advances in Neural Information Processing Systems},
  volume~32. Curran Associates, Inc.

\bibitem[{Du et~al.(2017)Du, Shao, and Cardie}]{du-etal-2017-learning}
Xinya Du, Junru Shao, and Claire Cardie. 2017.
\newblock \href {https://doi.org/10.18653/v1/P17-1123} {Learning to ask: Neural
  question generation for reading comprehension}.
\newblock In \emph{Proceedings of the 55th Annual Meeting of the Association
  for Computational Linguistics (Volume 1: Long Papers)}, pages 1342--1352,
  Vancouver, Canada. Association for Computational Linguistics.

\bibitem[{Fan et~al.(2018)Fan, Grangier, and Auli}]{fan-etal-2018-controllable}
Angela Fan, David Grangier, and Michael Auli. 2018.
\newblock \href {https://doi.org/10.18653/v1/W18-2706} {Controllable
  abstractive summarization}.
\newblock In \emph{Proceedings of the 2nd Workshop on Neural Machine
  Translation and Generation}, pages 45--54, Melbourne, Australia. Association
  for Computational Linguistics.

\bibitem[{Fan et~al.(2019)Fan, Jernite, Perez, Grangier, Weston, and
  Auli}]{fan-etal-2019-eli5}
Angela Fan, Yacine Jernite, Ethan Perez, David Grangier, Jason Weston, and
  Michael Auli. 2019.
\newblock \href {https://doi.org/10.18653/v1/P19-1346} {{ELI}5: Long form
  question answering}.
\newblock In \emph{Proceedings of the 57th Annual Meeting of the Association
  for Computational Linguistics}, pages 3558--3567, Florence, Italy.
  Association for Computational Linguistics.

\bibitem[{Ficler and Goldberg(2017)}]{ficler-goldberg-2017-controlling}
Jessica Ficler and Yoav Goldberg. 2017.
\newblock \href {https://doi.org/10.18653/v1/W17-4912} {Controlling linguistic
  style aspects in neural language generation}.
\newblock In \emph{Proceedings of the Workshop on Stylistic Variation}, pages
  94--104, Copenhagen, Denmark. Association for Computational Linguistics.

\bibitem[{FitzGerald et~al.(2018)FitzGerald, Michael, He, and
  Zettlemoyer}]{fitzgerald-etal-2018-large}
Nicholas FitzGerald, Julian Michael, Luheng He, and Luke Zettlemoyer. 2018.
\newblock \href {https://doi.org/10.18653/v1/P18-1191} {Large-scale {QA}-{SRL}
  parsing}.
\newblock In \emph{Proceedings of the 56th Annual Meeting of the Association
  for Computational Linguistics (Volume 1: Long Papers)}, pages 2051--2060,
  Melbourne, Australia. Association for Computational Linguistics.

\bibitem[{Flor and Riordan(2018)}]{flor-riordan-2018-semantic}
Michael Flor and Brian Riordan. 2018.
\newblock \href {https://doi.org/10.18653/v1/W18-0530} {A semantic role-based
  approach to open-domain automatic question generation}.
\newblock In \emph{Proceedings of the Thirteenth Workshop on Innovative Use of
  {NLP} for Building Educational Applications}, pages 254--263, New Orleans,
  Louisiana. Association for Computational Linguistics.

\bibitem[{Gao et~al.(2019)Gao, Li, King, and
  Lyu}]{gao-etal-2019-interconnected}
Yifan Gao, Piji Li, Irwin King, and Michael~R. Lyu. 2019.
\newblock \href {https://doi.org/10.18653/v1/P19-1480} {Interconnected question
  generation with coreference alignment and conversation flow modeling}.
\newblock In \emph{Proceedings of the 57th Annual Meeting of the Association
  for Computational Linguistics}, pages 4853--4862, Florence, Italy.
  Association for Computational Linguistics.

\bibitem[{Harman and Liberman(1993)}]{AP_tipster}
Donna Harman and Mark Liberman. 1993.
\newblock {TIPSTER} complete {LDC93T3A}. web download. {P}hiladelphia:
  {L}inguistic {D}ata {C}onsortium.

\bibitem[{Heilman and Smith(2010)}]{heilman-smith-2010-good}
Michael Heilman and Noah~A. Smith. 2010.
\newblock \href {https://aclanthology.org/N10-1086} {Good question! statistical
  ranking for question generation}.
\newblock In \emph{Human Language Technologies: The 2010 Annual Conference of
  the North {A}merican Chapter of the Association for Computational
  Linguistics}, pages 609--617, Los Angeles, California. Association for
  Computational Linguistics.

\bibitem[{Hu et~al.(2017)Hu, Yang, Liang, Salakhutdinov, and
  Xing}]{hu2017toward}
Zhiting Hu, Zichao Yang, Xiaodan Liang, Ruslan Salakhutdinov, and Eric~P. Xing.
  2017.
\newblock \href {http://proceedings.mlr.press/v70/hu17e.html} {Toward
  controlled generation of text}.
\newblock In \emph{Proceedings of the 34th International Conference on Machine
  Learning, {ICML} 2017, Sydney, NSW, Australia, 6-11 August 2017}, volume~70
  of \emph{Proceedings of Machine Learning Research}, pages 1587--1596. {PMLR}.

\bibitem[{Joachims et~al.(2007)Joachims, Li, Liu, and
  Zhai}]{joachims2007learning}
Thorsten Joachims, Hang Li, Tie{-}Yan Liu, and ChengXiang Zhai. 2007.
\newblock \href {https://doi.org/10.1145/1328964.1328974} {Learning to rank for
  information retrieval {(LR4IR} 2007)}.
\newblock \emph{{SIGIR} Forum}, 41(2):58--62.

\bibitem[{Keskar et~al.(2019)Keskar, McCann, Varshney, Xiong, and
  Socher}]{keskar2019ctrl}
Nitish~Shirish Keskar, Bryan McCann, Lav~R. Varshney, Caiming Xiong, and
  Richard Socher. 2019.
\newblock \href {http://arxiv.org/abs/1909.05858} {{CTRL:} {A} conditional
  transformer language model for controllable generation}.
\newblock \emph{CoRR}, abs/1909.05858.

\bibitem[{Kikuchi et~al.(2016)Kikuchi, Neubig, Sasano, Takamura, and
  Okumura}]{kikuchi-etal-2016-controlling}
Yuta Kikuchi, Graham Neubig, Ryohei Sasano, Hiroya Takamura, and Manabu
  Okumura. 2016.
\newblock \href {https://doi.org/10.18653/v1/D16-1140} {Controlling output
  length in neural encoder-decoders}.
\newblock In \emph{Proceedings of the 2016 Conference on Empirical Methods in
  Natural Language Processing}, pages 1328--1338, Austin, Texas. Association
  for Computational Linguistics.

\bibitem[{Kingma and Ba(2015)}]{DBLP:journals/corr/KingmaB14}
Diederik~P. Kingma and Jimmy Ba. 2015.
\newblock \href {http://arxiv.org/abs/1412.6980} {Adam: {A} method for
  stochastic optimization}.
\newblock In \emph{3rd International Conference on Learning Representations,
  {ICLR} 2015, San Diego, CA, USA, May 7-9, 2015, Conference Track
  Proceedings}.

\bibitem[{Ko et~al.(2020)Ko, Chen, Huang, Durrett, and Li}]{ko2020inquisitive}
Wei{-}Jen Ko, Te{-}Yuan Chen, Yiyan Huang, Greg Durrett, and Junyi~Jessy Li.
  2020.
\newblock \href {https://doi.org/10.18653/v1/2020.emnlp-main.530} {Inquisitive
  question generation for high level text comprehension}.
\newblock In \emph{Proceedings of the 2020 Conference on Empirical Methods in
  Natural Language Processing, {EMNLP} 2020, Online, November 16-20, 2020},
  pages 6544--6555. Association for Computational Linguistics.

\bibitem[{Krause et~al.(2021)Krause, Gotmare, McCann, Keskar, Joty, Socher, and
  Rajani}]{krause-etal-2021-gedi-generative}
Ben Krause, Akhilesh~Deepak Gotmare, Bryan McCann, Nitish~Shirish Keskar,
  Shafiq Joty, Richard Socher, and Nazneen~Fatema Rajani. 2021.
\newblock \href {https://doi.org/10.18653/v1/2021.findings-emnlp.424}
  {{G}e{D}i: Generative discriminator guided sequence generation}.
\newblock In \emph{Findings of the Association for Computational Linguistics:
  EMNLP 2021}, pages 4929--4952, Punta Cana, Dominican Republic. Association
  for Computational Linguistics.

\bibitem[{Kumar and Black(2020)}]{kumar2020clarq}
Vaibhav Kumar and Alan~W. Black. 2020.
\newblock \href {https://doi.org/10.18653/v1/2020.acl-main.651} {Clarq: {A}
  large-scale and diverse dataset for clarification question generation}.
\newblock In \emph{Proceedings of the 58th Annual Meeting of the Association
  for Computational Linguistics, {ACL} 2020, Online, July 5-10, 2020}, pages
  7296--7301. Association for Computational Linguistics.

\bibitem[{Labutov et~al.(2015)Labutov, Basu, and Vanderwende}]{labutov2015deep}
Igor Labutov, Sumit Basu, and Lucy Vanderwende. 2015.
\newblock \href {https://doi.org/10.3115/v1/P15-1086} {Deep questions without
  deep understanding}.
\newblock In \emph{Proceedings of the 53rd Annual Meeting of the Association
  for Computational Linguistics and the 7th International Joint Conference on
  Natural Language Processing (Volume 1: Long Papers)}, pages 889--898,
  Beijing, China. Association for Computational Linguistics.

\bibitem[{Lewis et~al.(2020)Lewis, Liu, Goyal, Ghazvininejad, Mohamed, Levy,
  Stoyanov, and Zettlemoyer}]{lewis-etal-2020-bart}
Mike Lewis, Yinhan Liu, Naman Goyal, Marjan Ghazvininejad, Abdelrahman Mohamed,
  Omer Levy, Veselin Stoyanov, and Luke Zettlemoyer. 2020.
\newblock \href {https://doi.org/10.18653/v1/2020.acl-main.703} {{BART}:
  Denoising sequence-to-sequence pre-training for natural language generation,
  translation, and comprehension}.
\newblock In \emph{Proceedings of the 58th Annual Meeting of the Association
  for Computational Linguistics}, pages 7871--7880, Online. Association for
  Computational Linguistics.

\bibitem[{Lin(2004)}]{lin-2004-rouge}
Chin-Yew Lin. 2004.
\newblock \href {https://aclanthology.org/W04-1013} {{ROUGE}: A package for
  automatic evaluation of summaries}.
\newblock In \emph{Text Summarization Branches Out}, pages 74--81, Barcelona,
  Spain. Association for Computational Linguistics.

\bibitem[{Liu et~al.(2009)}]{liu2009learning}
Tie-Yan Liu et~al. 2009.
\newblock Learning to rank for information retrieval.
\newblock \emph{Foundations and Trends{\textregistered} in Information
  Retrieval}, 3(3):225--331.

\bibitem[{Liu et~al.(2019)Liu, Ott, Goyal, Du, Joshi, Chen, Levy, Lewis,
  Zettlemoyer, and Stoyanov}]{liu2019roberta}
Yinhan Liu, Myle Ott, Naman Goyal, Jingfei Du, Mandar Joshi, Danqi Chen, Omer
  Levy, Mike Lewis, Luke Zettlemoyer, and Veselin Stoyanov. 2019.
\newblock \href {http://arxiv.org/abs/1907.11692} {Roberta: {A} robustly
  optimized {BERT} pretraining approach}.
\newblock \emph{CoRR}, abs/1907.11692.

\bibitem[{Mann and Thompson(1988)}]{mann1988rhetorical}
William~C Mann and Sandra~A Thompson. 1988.
\newblock Rhetorical structure theory: Toward a functional theory of text
  organization.
\newblock \emph{Text-interdisciplinary Journal for the Study of Discourse},
  8(3):243--281.

\bibitem[{Marcus et~al.(1993)Marcus, Santorini, and
  Marcinkiewicz}]{marcus-etal-1993-building}
Mitchell~P. Marcus, Beatrice Santorini, and Mary~Ann Marcinkiewicz. 1993.
\newblock \href {https://aclanthology.org/J93-2004} {Building a large annotated
  corpus of {E}nglish: The {P}enn {T}reebank}.
\newblock \emph{Computational Linguistics}, 19(2):313--330.

\bibitem[{Mitkov and Ha(2003)}]{mitkov-ha-2003-computer}
Ruslan Mitkov and Le~An Ha. 2003.
\newblock \href {https://aclanthology.org/W03-0203} {Computer-aided generation
  of multiple-choice tests}.
\newblock In \emph{Proceedings of the {HLT}-{NAACL} 03 Workshop on Building
  Educational Applications Using Natural Language Processing}, pages 17--22.

\bibitem[{Nakanishi et~al.(2019)Nakanishi, Kobayashi, and
  Hayashi}]{nakanishi2019towards}
Mao Nakanishi, Tetsunori Kobayashi, and Yoshihiko Hayashi. 2019.
\newblock \href {https://doi.org/10.18653/v1/D19-5809} {Towards answer-unaware
  conversational question generation}.
\newblock In \emph{Proceedings of the 2nd Workshop on Machine Reading for
  Question Answering}, pages 63--71, Hong Kong, China. Association for
  Computational Linguistics.

\bibitem[{Olney et~al.(2012)Olney, Graesser, and
  Person}]{DBLP:journals/dad/OlneyGP12}
Andrew~McGregor Olney, Arthur~C. Graesser, and Natalie~K. Person. 2012.
\newblock \href {http://dad.uni-bielefeld.de/index.php/dad/article/view/1480}
  {Question generation from concept maps}.
\newblock \emph{Dialogue Discourse}, 3(2):75--99.

\bibitem[{Ott et~al.(2019)Ott, Edunov, Baevski, Fan, Gross, Ng, Grangier, and
  Auli}]{ott2019fairseq}
Myle Ott, Sergey Edunov, Alexei Baevski, Angela Fan, Sam Gross, Nathan Ng,
  David Grangier, and Michael Auli. 2019.
\newblock \href {https://doi.org/10.18653/v1/N19-4009} {fairseq: A fast,
  extensible toolkit for sequence modeling}.
\newblock In \emph{Proceedings of the 2019 Conference of the North {A}merican
  Chapter of the Association for Computational Linguistics (Demonstrations)},
  pages 48--53, Minneapolis, Minnesota. Association for Computational
  Linguistics.

\bibitem[{Papineni et~al.(2002)Papineni, Roukos, Ward, and
  Zhu}]{papineni-etal-2002-bleu}
Kishore Papineni, Salim Roukos, Todd Ward, and Wei-Jing Zhu. 2002.
\newblock \href {https://doi.org/10.3115/1073083.1073135} {{B}leu: a method for
  automatic evaluation of machine translation}.
\newblock In \emph{Proceedings of the 40th Annual Meeting of the Association
  for Computational Linguistics}, pages 311--318, Philadelphia, Pennsylvania,
  USA. Association for Computational Linguistics.

\bibitem[{Pascual et~al.(2021)Pascual, Egressy, Meister, Cotterell, and
  Wattenhofer}]{pascual-etal-2021-plug-play}
Damian Pascual, Beni Egressy, Clara Meister, Ryan Cotterell, and Roger
  Wattenhofer. 2021.
\newblock \href {https://doi.org/10.18653/v1/2021.findings-emnlp.334} {A
  plug-and-play method for controlled text generation}.
\newblock In \emph{Findings of the Association for Computational Linguistics:
  EMNLP 2021}, pages 3973--3997, Punta Cana, Dominican Republic. Association
  for Computational Linguistics.

\bibitem[{Pyatkin et~al.(2021)Pyatkin, Roit, Michael, Goldberg, Tsarfaty, and
  Dagan}]{pyatkin-etal-2021-asking}
Valentina Pyatkin, Paul Roit, Julian Michael, Yoav Goldberg, Reut Tsarfaty, and
  Ido Dagan. 2021.
\newblock \href {https://doi.org/10.18653/v1/2021.emnlp-main.108} {Asking it
  all: Generating contextualized questions for any semantic role}.
\newblock In \emph{Proceedings of the 2021 Conference on Empirical Methods in
  Natural Language Processing}, pages 1429--1441, Online and Punta Cana,
  Dominican Republic. Association for Computational Linguistics.

\bibitem[{Radford et~al.(2019)Radford, Wu, Child, Luan, Amodei, and
  Sutskever}]{radford2019language}
Alec Radford, Jeff Wu, Rewon Child, David Luan, Dario Amodei, and Ilya
  Sutskever. 2019.
\newblock Language models are unsupervised multitask learners.

\bibitem[{Rajpurkar et~al.(2016)Rajpurkar, Zhang, Lopyrev, and
  Liang}]{rajpurkar-etal-2016-squad}
Pranav Rajpurkar, Jian Zhang, Konstantin Lopyrev, and Percy Liang. 2016.
\newblock \href {https://doi.org/10.18653/v1/D16-1264} {{SQ}u{AD}: 100,000+
  questions for machine comprehension of text}.
\newblock In \emph{Proceedings of the 2016 Conference on Empirical Methods in
  Natural Language Processing}, pages 2383--2392, Austin, Texas. Association
  for Computational Linguistics.

\bibitem[{Roller et~al.(2021)Roller, Dinan, Goyal, Ju, Williamson, Liu, Xu,
  Ott, Smith, Boureau, and Weston}]{roller2020recipes}
Stephen Roller, Emily Dinan, Naman Goyal, Da~Ju, Mary Williamson, Yinhan Liu,
  Jing Xu, Myle Ott, Eric~Michael Smith, Y{-}Lan Boureau, and Jason Weston.
  2021.
\newblock \href {https://doi.org/10.18653/v1/2021.eacl-main.24} {Recipes for
  building an open-domain chatbot}.
\newblock In \emph{Proceedings of the 16th Conference of the European Chapter
  of the Association for Computational Linguistics: Main Volume, {EACL} 2021,
  Online, April 19 - 23, 2021}, pages 300--325. Association for Computational
  Linguistics.

\bibitem[{Schiller et~al.(2021)Schiller, Daxenberger, and
  Gurevych}]{schiller-etal-2021-aspect}
Benjamin Schiller, Johannes Daxenberger, and Iryna Gurevych. 2021.
\newblock \href {https://doi.org/10.18653/v1/2021.naacl-main.34}
  {Aspect-controlled neural argument generation}.
\newblock In \emph{Proceedings of the 2021 Conference of the North American
  Chapter of the Association for Computational Linguistics: Human Language
  Technologies}, pages 380--396, Online. Association for Computational
  Linguistics.

\bibitem[{Scialom and Staiano(2020)}]{scialom-staiano-2020-ask}
Thomas Scialom and Jacopo Staiano. 2020.
\newblock \href {https://doi.org/10.18653/v1/2020.coling-main.202} {Ask to
  learn: A study on curiosity-driven question generation}.
\newblock In \emph{Proceedings of the 28th International Conference on
  Computational Linguistics}, pages 2224--2235, Barcelona, Spain (Online).
  International Committee on Computational Linguistics.

\bibitem[{Stasaski et~al.(2021)Stasaski, Rathod, Tu, Xiao, and
  Hearst}]{stasaski-etal-2021-automatically}
Katherine Stasaski, Manav Rathod, Tony Tu, Yunfang Xiao, and Marti~A. Hearst.
  2021.
\newblock \href {https://aclanthology.org/2021.bea-1.17} {Automatically
  generating cause-and-effect questions from passages}.
\newblock In \emph{Proceedings of the 16th Workshop on Innovative Use of NLP
  for Building Educational Applications}, pages 158--170, Online. Association
  for Computational Linguistics.

\bibitem[{Sultan et~al.(2020)Sultan, Chandel, Fernandez~Astudillo, and
  Castelli}]{sultan-etal-2020-importance}
Md~Arafat Sultan, Shubham Chandel, Ram{\'o}n Fernandez~Astudillo, and Vittorio
  Castelli. 2020.
\newblock \href {https://doi.org/10.18653/v1/2020.acl-main.500} {On the
  importance of diversity in question generation for {QA}}.
\newblock In \emph{Proceedings of the 58th Annual Meeting of the Association
  for Computational Linguistics}, pages 5651--5656, Online. Association for
  Computational Linguistics.

\bibitem[{Sun et~al.(2018)Sun, Liu, Lyu, He, Ma, and
  Wang}]{sun-etal-2018-answer}
Xingwu Sun, Jing Liu, Yajuan Lyu, Wei He, Yanjun Ma, and Shi Wang. 2018.
\newblock \href {https://doi.org/10.18653/v1/D18-1427} {Answer-focused and
  position-aware neural question generation}.
\newblock In \emph{Proceedings of the 2018 Conference on Empirical Methods in
  Natural Language Processing}, pages 3930--3939, Brussels, Belgium.
  Association for Computational Linguistics.

\bibitem[{Sutskever et~al.(2014)Sutskever, Vinyals, and
  Le}]{sutskever2014sequence}
Ilya Sutskever, Oriol Vinyals, and Quoc~V. Le. 2014.
\newblock \href
  {https://proceedings.neurips.cc/paper/2014/hash/a14ac55a4f27472c5d894ec1c3c743d2-Abstract.html}
  {Sequence to sequence learning with neural networks}.
\newblock In \emph{Advances in Neural Information Processing Systems 27: Annual
  Conference on Neural Information Processing Systems 2014, December 8-13 2014,
  Montreal, Quebec, Canada}, pages 3104--3112.

\bibitem[{Tsai et~al.(2021)Tsai, Oraby, Perera, Kao, Du, Narayan{-}Chen, Chung,
  and Hakkani{-}T{\"{u}}r}]{tsai2021style}
Alicia~Y. Tsai, Shereen Oraby, Vittorio Perera, Jiun{-}Yu Kao, Yuheng Du,
  Anjali Narayan{-}Chen, Tagyoung Chung, and Dilek Hakkani{-}T{\"{u}}r. 2021.
\newblock \href {http://arxiv.org/abs/2109.12211} {Style control for
  schema-guided natural language generation}.
\newblock \emph{CoRR}, abs/2109.12211.

\bibitem[{Tu et~al.(2019)Tu, Ding, Yu, and Gimpel}]{tu-etal-2019-generating}
Lifu Tu, Xiaoan Ding, Dong Yu, and Kevin Gimpel. 2019.
\newblock \href {https://doi.org/10.18653/v1/D19-5605} {Generating diverse
  story continuations with controllable semantics}.
\newblock In \emph{Proceedings of the 3rd Workshop on Neural Generation and
  Translation}, pages 44--58, Hong Kong. Association for Computational
  Linguistics.

\bibitem[{Vaswani et~al.(2017)Vaswani, Shazeer, Parmar, Uszkoreit, Jones,
  Gomez, Kaiser, and Polosukhin}]{DBLP:conf/nips/VaswaniSPUJGKP17}
Ashish Vaswani, Noam Shazeer, Niki Parmar, Jakob Uszkoreit, Llion Jones,
  Aidan~N. Gomez, Lukasz Kaiser, and Illia Polosukhin. 2017.
\newblock \href
  {https://proceedings.neurips.cc/paper/2017/hash/3f5ee243547dee91fbd053c1c4a845aa-Abstract.html}
  {Attention is all you need}.
\newblock In \emph{Advances in Neural Information Processing Systems 30: Annual
  Conference on Neural Information Processing Systems 2017, December 4-9, 2017,
  Long Beach, CA, {USA}}, pages 5998--6008.

\bibitem[{Wang et~al.(2019)Wang, Wei, Fan, Liu, and Huang}]{wang2019multi}
Siyuan Wang, Zhongyu Wei, Zhihao Fan, Yang Liu, and Xuanjing Huang. 2019.
\newblock \href {https://doi.org/10.1609/aaai.v33i01.33017168} {A multi-agent
  communication framework for question-worthy phrase extraction and question
  generation}.
\newblock In \emph{The Thirty-Third {AAAI} Conference on Artificial
  Intelligence, {AAAI} 2019, The Thirty-First Innovative Applications of
  Artificial Intelligence Conference, {IAAI} 2019, The Ninth {AAAI} Symposium
  on Educational Advances in Artificial Intelligence, {EAAI} 2019, Honolulu,
  Hawaii, USA, January 27 - February 1, 2019}, pages 7168--7175. {AAAI} Press.

\bibitem[{Wang et~al.(2020)Wang, Rao, Zhang, Qin, Tian, and
  Wang}]{wang-etal-2020-diversify}
Zhen Wang, Siwei Rao, Jie Zhang, Zhen Qin, Guangjian Tian, and Jun Wang. 2020.
\newblock \href {https://doi.org/10.18653/v1/2020.findings-emnlp.194}
  {Diversify question generation with continuous content selectors and question
  type modeling}.
\newblock In \emph{Findings of the Association for Computational Linguistics:
  EMNLP 2020}, pages 2134--2143, Online. Association for Computational
  Linguistics.

\bibitem[{Xu et~al.(2015)Xu, Callison{-}Burch, and
  Napoles}]{DBLP:journals/tacl/XuCN15}
Wei Xu, Chris Callison{-}Burch, and Courtney Napoles. 2015.
\newblock \href
  {https://tacl2013.cs.columbia.edu/ojs/index.php/tacl/article/view/549}
  {Problems in current text simplification research: New data can help}.
\newblock \emph{Trans. Assoc. Comput. Linguistics}, 3:283--297.

\bibitem[{Yates et~al.(2021)Yates, Nogueira, and Lin}]{lin2021pretrained}
Andrew Yates, Rodrigo Nogueira, and Jimmy Lin. 2021.
\newblock \href {https://doi.org/10.18653/v1/2021.naacl-tutorials.1}
  {Pretrained transformers for text ranking: {BERT} and beyond}.
\newblock In \emph{Proceedings of the 2021 Conference of the North American
  Chapter of the Association for Computational Linguistics: Human Language
  Technologies: Tutorials}, pages 1--4, Online. Association for Computational
  Linguistics.

\bibitem[{Zhang et~al.(2020)Zhang, Kishore, Wu, Weinberger, and
  Artzi}]{DBLP:conf/iclr/ZhangKWWA20}
Tianyi Zhang, Varsha Kishore, Felix Wu, Kilian~Q. Weinberger, and Yoav Artzi.
  2020.
\newblock \href {https://openreview.net/forum?id=SkeHuCVFDr} {Bertscore:
  Evaluating text generation with {BERT}}.
\newblock In \emph{8th International Conference on Learning Representations,
  {ICLR} 2020, Addis Ababa, Ethiopia, April 26-30, 2020}. OpenReview.net.

\bibitem[{Zhou et~al.(2019)Zhou, Zhang, and Wu}]{zhou-etal-2019-question}
Wenjie Zhou, Minghua Zhang, and Yunfang Wu. 2019.
\newblock \href {https://doi.org/10.18653/v1/D19-1622} {Question-type driven
  question generation}.
\newblock In \emph{Proceedings of the 2019 Conference on Empirical Methods in
  Natural Language Processing and the 9th International Joint Conference on
  Natural Language Processing (EMNLP-IJCNLP)}, pages 6032--6037, Hong Kong,
  China. Association for Computational Linguistics.

\end{thebibliography}
\bibliographystyle{acl_natbib}

\clearpage 
% \newpage

\appendix

\section{Appendix} \label{section:appendix}
\label{sec:appendix}
\begin{table*}[t!]
\centering
\small
\resizebox{\textwidth}{!}{
\begin{tabular}{lllllll}\toprule
Explanation & Elaboration & Background & Definition & Instantiation & Forward-looking & Other \\\midrule
why is(87) & what is(39) & who is(20) & what is(53) & what are(24) & how   will(5) & why is(4) \\
why did(75) & what are(31) & how long(19) & what are(17) & who are(16) & what will(2) & does this(3) \\
why was(51) & how did(27) & what was(17) & what does(15) & who is(8) & when will(2) & is it(2) \\
why are(50) & what does(17) & how did(16) & what do(6) & what other(5) & will the(2) & of which(2) \\
why were(24) & how is(12) & what is(14) & how is(2) & what kind(5) & what would(2) & yes what(1) \\
why would(21) & how do(12) & is that(10) & definition of(2) & what types(4) & what is(2) & what year(1) \\
why do(19) & how would(9) & how much(10) & is that(1) & which other(4) & what are(1) & does seaweed(1) \\
why does(18) & how are(9) & what are(9) & what 's(1) & what were(3) & how is(1) & does taxing(1) \\
why has(9) & how does(9) & is this(9) & does n't(1) & what sort(3) & were they(1) & what was(1) \\
what caused(7) & how many(8) & how many(8) & does note(1) & which companies(3) & is the(1) & this sounds(1) \\
what is(7) & what was(7) & is it(7) & does opportunities(1) & which year(3) & did they(1) & they must(1) \\
why will(7) & what kind(7) & are they(7) & i would(1) & what is(3) & how does(1) & there is(1) \\
is there(4) & how was(7) & where is(7) & does this(1) & which monday(2) & was their(1) & who is(1) \\
what makes(4) & what would(6) & how was(6) & what was(1) & what type(2) & did it(1) & what brass(1) \\
why have(3) & how much(6) & what did(6) & it means(1) & which officials(2) & so that(1) & should they(1) \\
what was(3) & how will(5) & how does(5) & who were(1) & in which(2) & how would(1) & which year(1) \\
why should(3) & how long(5) & where did(5) & what comprises(1) & who were(2) & what happened(1) & how many(1) \\
what were(2) & what makes(5) & what do(5) & thrift industry(1) & which countries(2) & would there(1) & is this(1) \\
why only(2) & how were(4) & why did(5) & how do(1) & which scientists(2) & would it(1) & is he(1) \\
why could(2) & in what(4) & when did(5) & who are(1) & which states(2) & will it(1) & which dollar(1)\\
\bottomrule
\end{tabular}
}
\caption{\label{ngram_annot_full} 
Most common leading bigrams in annotated questions (lowercased) for each type (counts shown in parentheses). 
%\Kevinin{Lingyu, since this is supplementary and therefore we have more space, can you show the top 20 bigrams for each type?}\Lingyuin{I put top 20 bigrams here, changed tokenization method.}
}
\end{table*}

%\begin{table}[t!]
%\centering
%\small
%\begin{tabular}{llllll}\toprule
%model & \cs & \css & \cssnqt & \cssrankqt & \cssrqt \\\midrule
%\% Acc & 35.07 & 49.39 & 34.11 & 37.45 & 85.57
%\\\bottomrule
%\end{tabular}
%\caption{\label{test_predict_acc} 
%Test accuracy for question type prediction for different models. 
%}
%\end{table}

\subsection{Experimental Setup} \label{subsection:expparameters}
For \cs, \css, \cssnqt, \cssrankqt, and \cssrqt, we use BART-large with the same settings. We train 15 epochs in total, using cross entropy loss (label smoothing with $\alpha = 0.1$), and set the maximum number of tokens per batch as 1024. There's a normalization layer after the embedding layer, and the embedding matrices for encoder input, decoder input, and decoder output are tied. For training, we use the Adam optimizer \citep{DBLP:journals/corr/KingmaB14} with learning rate 3e-5, weight decay 0.01, clip norm 0.1, dropout 0.1, and warm-up updates 500. 

For the question type classifier, we finetune RoBERTa-large for 15 epochs with batch size 8. We use Adam with learning rate 1e-5, weight decay 0.1, dropout 0.1, and warm-up updates 157. We use the same settings for the inquisitive vs.~informative classifier and pairwise ranking classifier except some hyperparameters. For the inquisitive vs.~informative classifier, we train for 10 epochs with batch size 32 and warm-up updates set to 300. For the pairwise ranking classifier, we train for 20 epochs with warm-up updates set to 387. Under this setting, we compute all warm-up updates with $6\%N_{tr}N_{epo}/{N_{bsz}}$, where $N_{tr}$ is the training set size, $N_{epo}$ is the number of training epochs, and $N_{bsz}$ is the batch size.
% \Lingyuin{warm-up updates for RoBERTa is odd because I choose it as 6\% of the total updates. For BART I didn't do that.}

\subsection{Leading Bigrams for Question Types}
Table~\ref{ngram_annot_full} shows the most common leading bigrams for each question type in our annotated data. We observe that for Background questions that start with ``what'', the bigrams are more scattered with multiple combinations, and ``how is/are/was/were/do'' etc.~appear more often in Elaboration than in Background questions.

\begin{small}
{
\begin{algorithm}[h]
\caption{Data selection for pairwise ranker}\label{data_select}
\textbf{Input}: $Q = \{q_{\mathit{rel}},\ q_{\mathit{nrel}}\}$, where $Q$ is the total set of generated questions for an instance, $q_{\mathit{rel}}$ is the set of relevant questions where $q_{\mathit{rel}} = \{(r_1, q_1), \cdots, (r_n, q_n)\}$, $q_{\mathit{nrel}}$ is the set of non-relevant questions, and $r_j$ is the rank for question $q_j$.
\begin{algorithmic}[1]
%\Procedure{$\mathit{Method}_1$}{$Q$}
 \LeftComment{Find relevant vs.~non-relevant}
\For{$q_j \in q_{\mathit{rel}}$} 
    \For{$q_k \in q_{\mathit{nrel}}$}
        \State \Yield $(q_j, q_k)$
    \EndFor
\EndFor
%\EndProcedure
%\Procedure{$\mathit{Method}_2$}{$q_{\mathit{rel}}$}
\\\LeftComment{Find questions with rank difference $\geq 2$}
\For{$j = 1, \cdots, n $}
    \State $k \gets j + 2$
    \While{$k \leq n $}
        \If{$r_{k} - r_{j} \geq 2$}
            \State \Yield $(q_{j}, q_{k})$
        \EndIf
        \State $k \gets k + 1$
    \EndWhile
\EndFor
%\EndProcedure
\end{algorithmic}
\end{algorithm}
}
\end{small}

% \begin{small}
% {
% \begin{algorithm}[ht]
% \caption{$Data\ Selection$}\label{data_select}
% \textbf{Input}: $\mathcal{X} = \sum_{i=1}^N\{s_i, (r_{1i}, q_{t_{1i}}), \cdots, (r_{ni}, q_{t_{ni}})\}$, where $N$ is the total number of annotated data, $s_i$ is the $i_{th}$ source sentence, $n = f(i)$ is the number of selected questions that varied by samples, $r_{ji}$ is the rank for question $q_{t_{ji}}$ (question type $t_{ji}$) and sorted in ascending order. Question type pool $p$ = \{Explanation, Elaboration, Background, Definition, Instantiation, Forward-looking\} includes all question types excluding `Others'.
% \begin{algorithmic}[1]
% % \Procedure{$get\_data$}{$\mathcal{X}$}
% \For{$i = 1, \cdots, N$} 
%     \State $p \gets [C, E, B, D, I, F]$
%     \For{$j = 1, \cdots, f(i)$}
%         \If{$t_{ji} \in p $} 
%             \State $p.remove(t_{ji})$
%         \EndIf
%         \State $k \gets j + 1$
%         \While{$k \leq f(i) $}
%             \Repeat
%                 \If{$t_{ki} \in p $} 
%                     \State $p.remove(t_{ki})$
%                 \EndIf
%                 \State $k \gets k + 1$
%             \Until {$r_{ji} + 2 > r_{ki}$}
%         \EndWhile
%         \For{$t \in p$}
%             \State \Yield $(s_i, q_{t_{ji}}, q_{t})$
%         \EndFor
%     \EndFor
% \EndFor
% % \EndProcedure
% \end{algorithmic}
% \end{algorithm}
% }
% \end{small}

\subsection{Data Selection for Pairwise Ranking Classifier}

Annotators may make the same or completely different choices, and two examples of annotator's ranking choices are shown in Table~\ref{table:rankexamples}. 
% \Lingyuin{Add examples here.}
\begin{table*}[t!]
\centering
\small
\begin{tabular}{@{}p{2cm}p{6.5cm}p{6.5cm}@{}}
\toprule
 {[}\emph{context}{]}{[}\emph{source} with span in \textbf{bold}{]}
 & 
 [NO\_CONTEXT][MILWAUKEE-The electric barrier on the \textbf{Chicago Sanitary and Ship Canal} that is considered the last line of defense to stop an Asian carp invasion of Lake Michigan has a problem : Fish can swim through it.] 
 & 
 [LOS ANGELES-Little-known fact : When it comes to extracting oxygen from the air we breathe, we humans are just OK. Birds are more efficient breathers than we are. So are alligators and, according to a new study, monitor lizards, and   probably most dinosaurs were as well.][Humans are what are called \textbf{tidal breathers}.] 
 \\\midrule
Definition & What is Chicago Sanitary and   Ship Canal? & what is a tidal breather? \\
Background & where is that? & Are they considered \textbackslash{}"tidal   breathers\textbackslash{}"? \\
Instantiation & Which section of the canal? & Who are these people? \\
Explanation & Why is this a problem? & Why are humans tidal breathers? \\
Forward & where is this? & How did they come up with this   term? \\
Elaboration & What is the name of the canal? & Are they not? \\
\midrule
Annotator A & 1. Forward 2. Explanation 3. Background & 1. Definition 2. Explanation 3. Forward \\
Annotator B & 1. Definition 2. Instantiation 3. Elaboration & 1. Definition 2. Explanation 3. Forward \\
\bottomrule
\end{tabular}
\caption{\label{table:rankexamples} Examples of different ranking choices of expert annotators.}
\end{table*}

\begin{table}[t!]
\centering
\small
\begin{tabular}{ll}\toprule
Question Type & \% Acc \\\midrule
Explanation & 97.82 \\
Elaboration & 65.84 \\
Background & 48.91 \\
Definition & 54.85 \\
Instantiation & 50.23 \\
Forward-looking & 0. \\\bottomrule
\end{tabular}
\caption{\label{control_acc} 
Test accuracy for question type prediction for model generation of different question types. 
}
\end{table}

Algorithm \ref{data_select} shows how training data is produced for the pairwise ranking classifier. The training instances are the 
combination of (a) a question $q_{\mathit{i}}$ from $q_{\mathit{rel}}$ and a question $q_{\mathit{j}}$ from $q_{\mathit{nrel}}$  and (line 2-6 in Algorithm \ref{data_select}) (b) two questions $q_{\mathit{i}}$ and $q_{\mathit{j}}$  from $q_{\mathit{rel}}$ if and only if the two questions are separated by $\ge$2 ranks (line 8-16 in Algorithm \ref{data_select}). 

\iffalse
\begin{figure}[t]
\centering
    % \begin{subfigure}{.45\textwidth}
    %     \centering
    %     \includegraphics[width=\linewidth]{figures/unigram_num.png}
    %     \caption{Number of question leading unigram.}
    % \end{subfigure}
    \begin{subfigure}{.45\textwidth}
        \centering
        \includegraphics[width=\linewidth]{figures/unigram_prob.png}
        % \caption{Distribution of question leading unigram (normalized over columns).}
    \end{subfigure}
\caption{\label{unigram_qtype} Distribution of question leading unigram (appear more than 10 times after lowercased) over annotated question types, normalized over columns. Frequency of the unigram on x axis decreases from left to right.}
\end{figure}

\begin{figure}[t]
\centering
    \begin{subfigure}{.45\textwidth}
        \centering
        \includegraphics[width=\linewidth]{figures/css.png}
        \caption{Context+source+span.}
    \end{subfigure}
    \begin{subfigure}{.45\textwidth}
        \centering
        \includegraphics[width=\linewidth]{figures/cssq.png}
        \caption{Context+source+span+question.}
    \end{subfigure}
\caption{\label{confusion_matrix} Heatmaps for confusion matrix of question type predictions. B: Background, C: Explanation (causal), D: Definition, E: Elaboration, F: Forward-looking, I: Instantiation, O: Others.}
\end{figure}
\fi

\begin{figure}[t]
\centering
    \begin{subfigure}{.45\textwidth}
        \centering
        \includegraphics[width=\linewidth]{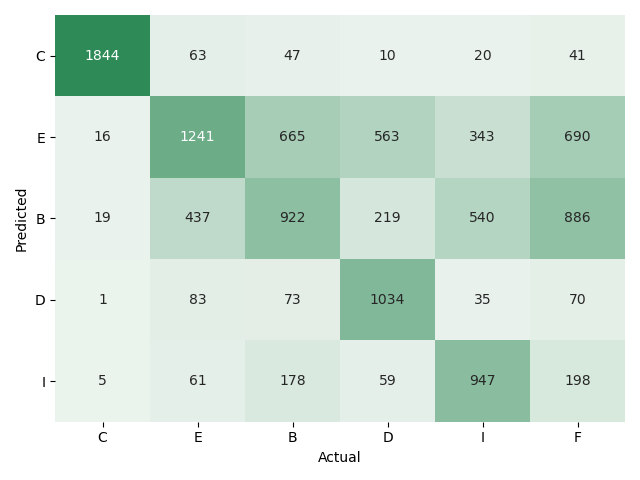}
    \end{subfigure}
\caption{\label{confusion_matrix} Heatmap showing confusion matrix for type controllability evaluation. The ``Actual'' type is the desired type passed as control code to the model, and the ``Predicted'' type is the output of running the question type classifier on the generated question. C: Explanation (causal), E: Elaboration, B: Background, D: Definition, I: Instantiation, F: Forward-looking.}
\end{figure}

\subsection{Controllability Evaluation}
% \Lingyuin{This is from context source span question model, only 1 run but I think should be enough for Appendix.}
We generate test set questions with six question types except ``Other'', and then classify the generations with our question type classifier.  The test accuracy is shown in  Table~\ref{control_acc}, with confusion matrix shown in Figure~\ref{confusion_matrix}. As the largest number in each row/column is along the diagonal (aside from forward-looking questions, which the classifier never predicts in this set), the model and classifier are in alignment a significant fraction of the time. We also observe that Explanation is doing well in both precision and recall, Elaboration and Background are tricky to discriminate from each other, and Definition and Instantiation are being classified with high precision though not with very high recall. When the model is asked to generate a forward-looking question, the classifier labels it as Elaboration or Background in most cases. This is likely because there are very few forward-looking questions in the training data. 

% When we classify generations of different models, the test accuracy is shown in Table~\ref{test_predict_acc}. While our oracle model \cssrqt scores highest, \css is the second highest, which implies that the \css could learn to ask questions of similar question types given gold spans.

\subsection{Additional Results}
All results in Table~\ref{traditional_metrics_full} and Table~\ref{Ko_metrics_full} are averaged over 5 different runs with standard deviations.

Table~\ref{traditional_metrics_full} reports BLEU scores for 1/2/3/4-grams. While \cs  always scores lowest and \cssrqt is always highest, \css is second-highest for BLEU-1, BLUE-2 and BLEU-3\footnote{The difference between \css and \cssrankqt  is too small in BLEU-3 to be shown in the table.}, and beat by \cssrankqt for BLEU-4.

Table~\ref{Ko_metrics_full} reports all the metric scores that are specifically implemented by \citet{ko2020inquisitive}. We see that \cssrankqt has lowest scores for \emph{Train-n}.
% and \emph{Train-3}, and second-lowest for \emph{Train-4} after \cssrqt. 
For \emph{Article-n}, the model order is changed when n is varied, e.g., \cs is higher than \cssrankqt on \emph{Article-1} but lower on \emph{Article-2} and \emph{Article-3}. Nevertheless, \cssnqt is always lower than \human on \emph{Article-n}, and other models are always higher than scores of \human. 
% \Lingyuin{The ko metrics are written refer to the main text. The comparison between models on Article-n is not preserved when n is varied.}
% \Kevinin{I'm confused.. if the comparison is not preserved when n is varied, shouldn't we change the text above that says that the comparison to human scores on Article-n is preserved for models? Or is that sentence saying the comparison is being preserved in some way other than as n is varied?}
% \Lingyuin{The model order is changed, e.g. \cs is higher than \cssrankqt on Article-2 but lower on Article-3. But \cssnqt and \cssrqt are always lower than human, and other models are always higher than human.}
% \Kevinin{yea I agree. can you change the incorrect text then to be what you wrote here in your comment?}

\begin{table*}[t!]
\centering
\small
\resizebox{\textwidth}{!}{
\begin{tabular}{cccccccccc}\toprule
Model & \%BLEU-1 & \%BLEU-2 & \%BLEU-3 & \%BLEU-4 & \%METEOR & \%ROUGE-L & \%F\textsubscript{BERT} & GPT2 ppl & Entropy \\\midrule
%context-source (\cs) 
\cs 
& 26.9\std{0.2} & 12.0\std{0.3} & 6.8\std{0.2} & 4.3\std{0.2} & 11.8\std{0.3} & 27.4\std{0.3} & 39.6\std{0.5} & 119\std{25} & 0.699\std{0.015} \\
%context-source-span (\css) 
\css 
& 35.1\std{0.9} & 19.4\std{0.5} & 12.4\std{0.4} & 8.5\std{0.4} & 17.5\std{0.7} & 36.1\std{0.5} & 47.6\std{0.4} & 148\std{10} & 0.726\std{0.062}  \\
%context-source-span-nqtype (\cssnqt) 
\cssnqt 
& 28.9\std{1.1} & 14.6\std{0.7} & 8.7\std{0.6} & 5.7\std{0.5} & 13.6\std{0.5} & 30.9\std{0.3} & 41.6\std{0.5} & 219\std{18} & 0.823\std{0.024}  \\
%context-source-span-rank (\cssrankqt) 
\cssrankqt 
& 33.4\std{1.4} & 18.9\std{1.0} & 12.4\std{0.8} & 8.6\std{0.6} & 18.3\std{0.4} & 35.3\std{0.7} & 47.4\std{0.8} & 89\std{7} & 0.612\std{0.025}  \\ 
%context-source-span-rqtype (\cssrqt) 
\cssrqt 
& 37.7\std{1.0} & 21.6\std{0.8} & 14.0\std{0.8} & 9.7\std{0.8} & 19.5\std{0.4} & 39.1\std{0.4} & 50.1\std{0.5} & 154\std{18} & 0.751\std{0.008}  \\
\bottomrule
\end{tabular}
}
\caption{\label{traditional_metrics_full} Automatic metrics on our test set for our models. 
%\Lingyuin{Didn't include Human because this table is for adding more BLEU scores}
}
\end{table*}

\begin{table*}[t!]
\centering
\small
\begin{tabular}{llllllll}\toprule
 & Train-2 & Train-3 & Train-4 & Article-1 & Article-2 & Article-3 & Span \\\midrule
%\citet{ko2020inquisitive} & 0.627 & 0.352 & 0.135 & 0.397 & 0.147 & 0.088 & 0.278 \\
Human & 0.467 & 0.203 & 0.059 & 0.386 & 0.126 & 0.064 & 0.354 \\\midrule
\cs & 0.518\std{0.018} & 0.267\std{0.019} & 0.097\std{0.009} & 0.469\std{0.020} & 0.186\std{0.020} & 0.104\std{0.018} & 0.184\std{0.007} \\
\css & 0.505\std{0.015} & 0.246\std{0.020} & 0.079\std{0.012} & 0.455\std{0.025} & 0.182\std{0.022} & 0.101\std{0.019} & 0.452\std{0.029} \\
\cssnqt & 0.530\std{0.006} & 0.288\std{0.012} & 0.102\std{0.012} & 0.315\std{0.015} & 0.090\std{0.010} & 0.041\std{0.006} & 0.346\std{0.023} \\
\cssrankqt & 0.473\std{0.013} & 0.218\std{0.015} & 0.068\std{0.010} & 0.445\std{0.018} & 0.195\std{0.016} & 0.112\std{0.013} & 0.542\std{0.030} \\
\cssrqt & 0.488\std{0.011} & 0.233\std{0.012} & 0.073\std{0.004} & 0.401\std{0.020} & 0.149\std{0.016} & 0.078\std{0.012} & 0.475\std{0.024}\\\bottomrule
% \multirow{2}{*}{Baselines} 
%  &\citep{ko2020inquisitive} & 0.627 & 0.352 & 0.135 & 0.397 & 0.147 & 0.088 & 0.278\\
%  & Human & 0.456 & 0.198 & 0.057 & 0.403 & 0.145 & 0.081 & 0.383 \\\hline % human scores are computed, not from paper, different from paper
% \multirow{8}{*}{w/o context} 
%  & source & 0.508 & 0.265 & 0.091 & 0.476 & 0.213 & 0.136 & 0.201 \\
%  & source-span & 0.498 & 0.256 & 0.085 & 0.466 & 0.201 & 0.120 & 0.490 \\
%  & source-nqtype & 0.547 & 0.321 & 0.111 & 0.303 & 0.066 & 0.027 & 0.111 \\
%  & source-span-nqtype & 0.540 & 0.303 & 0.114 & 0.318 & 0.091 & 0.042 & 0.346 \\
%  & source-rqtype & 0.502 & 0.256 & 0.087 & 0.375 & 0.120 & 0.057 & 0.203 \\
%  & source-span-rqtype & 0.484 & 0.235 & 0.075 & 0.402 & 0.145 & 0.076 & 0.470 \\
%  & source-rank & 0.482 & 0.238 & 0.078 & 0.412 & 0.161 & 0.091 & 0.222 \\
%  & source-span-rank & 0.469 & 0.220 & 0.070 & 0.449 & 0.194 & 0.112 & 0.543 \\
%   \hline
% \multirow{8}{*}{with context} 
%  & source & 0.521 & 0.274 & 0.101 & 0.469 & 0.201 & 0.119 & 0.190 \\
%  & source-span & 0.491 & 0.246 & 0.087 & 0.480 & 0.212 & 0.130 & 0.501 \\
%  & source-nqtype & 0.568 & 0.344 & 0.131 & 0.290 & 0.065 & 0.024 & 0.104 \\
%  & source-span-nqtype & 0.528 & 0.293 & 0.101 & 0.305 & 0.077 & 0.033 & 0.309 \\
%  & source-rqtype & 0.496 & 0.254 & 0.085 & 0.368 & 0.123 & 0.063 & 0.210 \\
%  & source-span-rqtype & 0.494 & 0.249 & 0.077 & 0.377 & 0.125 & 0.060 & 0.436 \\
%  & source-rank & 0.471 & 0.224 & 0.075 & 0.434 & 0.185 & 0.107 & 0.220 \\
%  & source-span-rank & 0.482 & 0.236 & 0.080 & 0.422 & 0.171 & 0.094 & 0.503 \\\hline
\end{tabular}
\caption{\label{Ko_metrics_full} Metric scores from \citet{ko2020inquisitive} that measure the extent of copying content from the training partition, articles, and spans in the source sentences to the generated questions. All scores are reported on our test set.}
\end{table*}

\subsection{Additional Examples}
Table~\ref{Qtype_distro_full} lists more annotated examples for each question type, and Table~\ref{table:wrongcssrank_full} includes examples (gold and generated questions by our models) that are judged ungrammatical by annotators.
\begin{table*}[t!]
%\begin{center}
\small
%\begin{tabularx}{\textwidth}{@{}l C{0.7} C{0.3}@{}}
%\begin{tabularx}{\textwidth}{@{} l X X @{}}
\begin{tabular}{@{}l p{8.4cm}p{4cm}@{}}
\toprule %\hline
\multirow{2}{*}{\makecell{Question Type \\(\# samples)}} & \multicolumn{2}{c}{Example} \\\cmidrule{2-3}%\cline{2-3} 
 &  [\emph{context}] [\emph{source} with span in \textbf{bold}] & Question \\ 
 \midrule %\hline
\multirow{1}{*}{Explanation (443)}
& [\dots Osip Nikiforov is recording Chopin's Etude Op. 10, No. 1, without capturing any of its sound.] [\textbf{Instead,} a sensor-equipped piano is recording the ``data'' of his performance \dots.] & Why is there a sensor-equipped piano recording data of his performance?\\\\
\multirow{2}{*}{Elaboration (364)} 
& [NO\_CONTEXT][Miami Shores, Fla., tech \textbf{consultant} Rudo Boothe, age 33, attributes his professional success \dots.] & For what company?\\\\
& [NO\_CONTEXT][The Agriculture Department says Americans seem to be eating a bit more each year but are \textbf{choosier} about what's on the menu.] & what are they choosing?\\\\
& [One of Ronald Reagan's attributes as President was that he rarely gave his blessing to the claptrap \dots.] [In fact, he liberated the U.S. from one of the world's most \textbf{corrupt} organizations -- UNESCO.] & How is UNESCO corrupt?\\\\
\multirow{1}{*}{Background (407)}
& [NO\_CONTEXT][\dots a young man and his mentor practice bullfighting techniques under the \textbf{light} of an atrium.] & Are they practicing at night?\\\\
\multirow{2}{*}{Definition (114)} 
& [NO\_CONTEXT][LOS ANGELES - The booming illegal international wildlife trade forced \textbf{conservationists} to do the unthinkable Tuesday \dots.] & Who were the conservationists?\\\\
& [People start their own businesses for many reasons. But a chance to fill out sales - tax records is rarely one of them.] [Red tape is the \textbf{bugaboo} of small business.] & what is a bugaboo?\\\\
\multirow{1}{*}{Instantiation (159)}
& [The Bush administration's nomination of Clarence Thomas to a seat on the federal appeals court here received a blow this week \dots] [People familiar with the Senate Judiciary Committee, \dots , said some \textbf{liberal members} of the panel are likely to question the ABA rating in hearings on the matter.] & Which liberal members are likely to question the ABA ratings?\\\\
\multirow{1}{*}{Forward-looking (31)} 
& [Bethlehem Steel Corp. has agreed in principle to form a joint venture with the world's second-largest steelmaker \dots.] [The entire division employs about \textbf{850 workers}.]& How will they need to increase or decrease staff?\\\\
\multirow{1}{*}{Other (32)}
& [\dots there's one easy way to make a July beach vacation even better than expected: Add seaweed \dots] [\dots his back covered in what looked like strands of \textbf{chartreuse cotton candy,} the 7-year-old Beijing boy was having the time of his life Sunday \dots] & Does seaweed look like cotton candy?\\
\bottomrule
%\end{tabularx}
\end{tabular}
%\end{center}
\caption{\label{Qtype_distro_full} Annotated question type distributions and salient examples of each question type. Context and source sentences are presented where the spans in source sentences are bold. 
}
\end{table*}
\begin{table*}[t!]
\centering
\small
\begin{tabular}{l|p{0.75\linewidth}}\toprule
\multirow{5}{*}{\human} 
 & why would it do that? \\
%  & 2 & what does she do now? \\
  & is it the aha? \\
  & in which year? \\
%  & 4 & why is he chosen? \\
%  & 5 & is it not easy? \\
%  & 6 & Why so many people year-over-year? \\
  & WHAT COUTRIES RECIEVED LOANS? \\
%  & 8 & Why do the Tamils not have a homeland, in their mind? \\
%  & 9 & is that cardio? \\
%  & 10 & Why don't even optimists expect progress? \\
%  & 11 & What time frame is this? \\
%  & 12 & How peacekeeping troops in Bosnia last month? \\
%  & 13 & Why do these companies not expect no long-term disruption in shipments   compare to the other companies in the region? \\
%  & 14 & Disarmament of who? \\
%  & 15 & Again, it doesn't let me highlight everything - just a single word. What   wrongdoing were the initially accused of? \\
%  & 16 & after what? \\
%  & 17 & How is Croatia likely to attack Serbs? \\
  & What specifically are the unhappy about with the direction? \\
%  & 19 & Why was Claes meeting Munich of Western defense leaders and military chiefs? \\
%  & 20 & how much has inflation? \\
%  & 21 & nativity happens for buddha? \\
%  & 22 & Expected by whom? \\
%  & 23 & How long does it take for her legs to turn to mush when they have this milk   fever? \\
%  & 24 & WHY DID INTEREST INCOME RISE? \\
%  & 25 & Which medical device companies are being worked with? \\
%  & 26 & When he decide? \\
%  & 27 & is it always routine? \\
%  & 28 & what kind of economic data? \\
%  & 29 & Why is there ability uncertain? \\
%  & 30 & Heat release from what? \\
%  & 31 & which person said it? \\
%  & 32 & are they likely to find them? \\
\midrule
\multirow{5}{*}{\cs} 
  & What goal does everyone have? \\
 & What happened that they didn't agree? \\
  & What kind of violence? \\
  & what are these signs? \\
  & What was Andrew Coltart doing at 69? \\
 \midrule

 \iffalse

 & 5 & Why was he added to the list? \\
 & 6 & Why is there no result to a political solution? \\
 & 7 & What Western governments? \\
 & 8 & What was the reasoning to call into question this idea? \\
 & 9 & Why are farmers using it to make better decisions? \\
 & 10 & Why was Japan in war? \\
 & 11 & What is an Example Prime Rate? \\
 & 12 & What is the persistence of scandal? \\
 & 13 & How were they excessivelyZealous? \\
 & 14 & What were these ventures, and what are they? \\
 & 15 & Why were they on the strength of their chemical business? \\
 & 16 & How do the bugs have mastered these mechanisms? \\
 & 17 & Why is he holdingover? \\
 & 18 & What would be the hurt? \\
 & 19 & What are the other businesses? \\
 & 20 & What are the authorities? \\
 & 21 & How did the Yugoslav's help to save the Jews? \\
 & 22 & Why did they decide tobenefit passengers? \\
 & 23 & How many years have been involved? \\
 & 24 & What are \textbackslash{}"small-minded rules\textbackslash{}"? \\
 & 25 & What are these boys tops for? \\
 & 26 & Why were children sent home when they were infestation with head lice? \\
 & 27 & How much did it profit in yearago period? \\
 & 28 & How does the lack of confidence in Washington right itself showing up?
 \fi
\multirow{5}{*}{\css}  & Why weren't the details unavailable? \\
  & Why is there a hard time posting an upset over Germany? \\
  & What is their goal in common? \\
  & Which lawmakers and others arguing? \\
  & How did they inflating the stock price? \\
 \midrule

  \iffalse

 & 5 & Who was Buddha's birth? \\
 & 6 & What is P.R.R.? \\
 & 7 & of which year? \\
 & 8 & Why is he waiting and what is he Waiting for? \\
 & 9 & How was the Kobe Earthquake devastating? \\
 & 10 & Why does it only want to sell substantially all of its other core businesses   if it has the right to regain the subsidiary if it is not selling   substantially \\
 & 11 & Why is it fifty years ago? \\
 & 12 & why did he struggled? \\
 & 13 & What happened in the Soviet Union's landers to land a rover on the moon? \\
 & 14 & how are his bullish on emerging growth stocks? \\
 & 15 & How did they reach this conclusions? \\
 & 16 & how soon can it be gotten on her feet? \\
 & 17 & Why did Varian Associates Inc. report a 95 percenst profit plunging? \\
 & 18 & What are some of their major clients? \\
 & 19 & How do you rapidly copying it so there's enough to identify what's causing   your illness? \\
 & 20 & Where do Blackburn and their fan go? \\
 & 21 & Which local medical device companies? \\
 & 22 & What is the percentage for motherhood in relation to wages? \\
 & 23 & Which world powers? \\
 & 24 & How does passe mean? \\
 & 25 & Which country is pessimists about? \\
 & 26 & How did the par- better players get in the top 30? \\
 & 27 & Why was Integrated Resources Equity Corp. renamedRoyal Alliance Associates   Inc.? \\
 \fi
\multirow{5}{*}{\cssnqt}  & which meetings? What meetings? \\
  & What are the details about this other than that? What details? \\
  & What goal? What goal? \\
 & what were they? \\
 & what prefecture? \\
 \midrule

\multirow{5}{*}{\cssrankqt}  & Who are the Serbs from Croatia and Bosnian Muslims opposed to the Bosnian   government? \\
  & Why would NATO take in Poland, Hungary and others asMembers? \\
  & How does Dominican authorities know the whereabouts of the banker and two   Dominicans? \\
  & How does a report about AIDS come to a conclusion? \\
  & Why is this symbol of America? \\
 \iffalse
 & 5 & Why are boys with tops and Frisbee tossers? \\
 & 6 & Why was this center located at that location? \\
 & 7 & Why is the VP of Public Affairs for the American Wind Energy Association? \\
 & 8 & How was the election of Heshy Bucholz so important? \\
 & 9 & What does Blackburn mean to Blackburn? \\
 & 10 & How did she get the claw over the side of the sailboat in order? \\
 & 11 & Why did net climb 19\%? \\
 & 12 & How did he admit? \\
 & 13 & Why is he not racing on oval tracks before? \\
 \fi
\midrule
\multirow{5}{*}{\cssrqt} & How many peacekeepers? \\
  & How was anreement to conceal the agreement made? \\
  & Did these talks involve a lot of talks? \\
  & How long has the explosion been taking place? \\
  & What are terms and syndicate manager? \\
\bottomrule
\end{tabular}
\caption{\label{table:wrongcssrank_full} Examples of gold questions from \inquisitive and questions generated by models that are judged as ungrammatical by annotators.}
\end{table*}

\end{document}